\pdfoutput=1
\PassOptionsToPackage{prologue,dvipsnames}{xcolor}
\documentclass[11pt]{article}

\usepackage[final]{acl}

\usepackage{times}
\usepackage{latexsym}

\usepackage[T1]{fontenc}

\usepackage[utf8]{inputenc}

\usepackage{microtype}

\usepackage{inconsolata}

\usepackage{caption}
\usepackage{graphicx}
\usepackage{float} 
\usepackage{subcaption}
\usepackage{amsmath} 
\usepackage{hyperref}
\usepackage{url}
\usepackage{wrapfig}
\usepackage{graphicx}
\usepackage{amsthm}
\newtheorem{definition}{Definition}[section]

\usepackage{soul}
\usepackage{multirow}
\usepackage{tcolorbox}
\usepackage{pifont}
\usepackage{makecell}
\usepackage{arydshln}
\usepackage{siunitx}
\usepackage{booktabs}

\usepackage{amsmath,amsfonts,bm}

\def\eqref#1{equation~\ref{#1}}

\def\1{\bm{1}}

\DeclareMathAlphabet{\mathsfit}{\encodingdefault}{\sfdefault}{m}{sl}
\SetMathAlphabet{\mathsfit}{bold}{\encodingdefault}{\sfdefault}{bx}{n}

\def\gA{{\mathcal{A}}}

\def\gD{{\mathcal{D}}}

\def\papertitle{{AIDE}}
\def\papertitles{{AIDE~}}

\newcommand{\attr}{a}

\usepackage{lipsum}
\tcbuselibrary{skins,breakable}
\usetikzlibrary{shadings,shadows}
\newenvironment{gblock}[1]{
\tcolorbox[beamer,
noparskip,
title=#1]}
{\endtcolorbox}

\title{AIDE: \underline{A}ttribute-Guided Mult\underline{I}-Hop \underline{D}ata \underline{E}xpansion \\ for Data Scarcity in Task-Specific Fine-tuning }

\author{
 \textbf{Jiayu Li\textsuperscript{1,3,\thanks{Work was done during an internship at AWS.}}},
 \textbf{Xuan Zhu\textsuperscript{2}},
 \textbf{Fang Liu\textsuperscript{2}},
 \textbf{Yanjun Qi\textsuperscript{2,3}}
\\
\\
 \textsuperscript{1}Syracuse University,
 \textsuperscript{2}AWS Bedrock Science
\\
 \small{
   \textbf{$^{3}$Correspondence:} \href{mailto:email@domain}{jli221@data.syr.edu, yanjunqi@amazon.com}
 }
}

\begin{document}
\maketitle
\begin{abstract}
Fine-tuning large language models (LLMs) for specific tasks requires diverse, high-quality training data. However, obtaining sufficient relevant data remains a significant challenge. Existing data synthesis methods either depend on extensive seed datasets or struggle to balance task relevance and data diversity. To address these challenges, we propose \textbf{A}ttribute-guided mult\textbf{I}-hop \textbf{D}ata \textbf{E}xpansion (\papertitle), a novel data synthesis framework that uses a multi-hop process to expand very few seed data points while ensuring data diversity and task relevance. \papertitles extracts the main topic and key knowledge attributes from the seeds to guide the synthesis steps. The process repeats for $K$ hops, using the generated data as seeds. To prevent irrelevant data generation as the hop depth increases, \papertitle~incorporates a residual connection mechanism. Our empirical results show that \papertitle~enables fine-tuning of Mistral-7B, Llama-3.1-8B and Llama-3.2-3B from 10 seeds, surpassing the models fine-tuned on human curated data. Furthermore, \papertitle~outperforms state-of-the-art data synthesis methods, such as Evol-Instruct, by over $30\%$ in task-specific fine-tuning. Code is available at \url{https://github.com/Code4Graph/AIDE}.
\end{abstract}

\section{Introduction}
Fine-tuning with task-specific training data is essential because it allows a pre-trained model to adapt and optimize for a specific task, resulting in better performance in that domain. However, task-specific data is insufficient or unavailable for many use cases, and manually curating the data is labor intensive~\cite{gandhi2024bettersyntheticdataretrieving}.

To overcome the limitation, an approach from~\cite{wei2022finetuned, xu-etal-2022-zeroprompt} samples task-specific training data from public NLP datasets, but the sampling often covers limited information. Another category of recent methods leverages the capabilities of LLMs to automatically generate large-scale synthetic data, enabling the training of advanced models in specific task domains. For example, Prompt2Model \citep{viswanathan-etal-2023-prompt2model} and DataTune \citep{gandhi2024bettersyntheticdataretrieving} rely on several candidate datasets to synthesize task-specific data for fine-tuning LLMs. However, these methods either require a large set of seed data for rewriting or produce synthetic data that lacks task relevance and diversity, as they do not maintain sufficient control over the synthesis process.

To address these challenges, we propose \textbf{\papertitle} (\textbf{A}ttribute-guided mult\textbf{I}-hop \textbf{D}ata \textbf{E}xpansion), a novel data synthesis framework that generates abundant training data from a small set of seed inputs, as shown in Figure \ref{overview}. Our framework focuses on maintaining high task relevance, diversity, and quality in the synthetic data for specific tasks. \papertitles uses LLMs as key players via a multi-hop synthesis process. Each hop in \papertitles begins by extracting the main topic and important knowledge attributes from a seed sample using a LLM. 
This builds knowledge triplets, and \papertitles traverses these triplets (each consisting of a topic, relationship, and attribute) to synthesize new data points.
In the next hop, each newly generated data point becomes a seed, and the process repeats until reaching a depth of $K$ hops. This multi-hop mechanism allows for recursive data synthesis along all paths of a process tree, enabling the generation of large-volume data from just a few seeds. Extracted attributes act as control nodes in the multi-hop tree, ensuring the generated data points remain relevant to the target task. We also introduce personas as new key attributes, enhancing the generation of diverse data. As the depth of the recursive synthesis increases, the relevance of the synthetic data may diminish. To address this, we propose a residual connection mechanism to reduce irrelevance.

To validate \papertitle, we conduct experiments with three pretrained models (Mistral-7B, Llama-3.1-8B, and Llama-3.2-3B). We evaluate the performance of these models when fine-tuned with synthetic data generated by \papertitle, comparing the results against models fine-tuned with human-curated (gold) data and synthetic data from state-of-the-art (SOTA) methods. Our evaluations span a range of tasks from well-known benchmarks, including industrial datasets like MedQA~\cite{jin2020disease} and FinBen~\cite{xie2024finbenholisticfinancialbenchmark}, as well as BIG-Bench~\cite{srivastava2023beyond}, MMLU~\cite{hendryckstest2021}, ARC-Challenge~\cite{Clark2018ThinkYH}, GSM8K~\cite{cobbe2021gsm8k}, and TruthfulQA~\cite{lin-etal-2022-truthfulqa}. For comparison, we include SOTA data synthesis methods such as Evol-Instruct~\cite{xu2024wizardlm}, DataTune~\cite{gandhi2024bettersyntheticdataretrieving}, and Prompt2Model~\cite{viswanathan-etal-2023-prompt2model}. Our main contributions are as follows:
\begin{itemize}
    \item We introduce \papertitle, a novel data synthesis framework that has a multi-hop synthesis, guided by attributes and personas, to generate abundant, task-relevant, diverse, and high-quality data from only a few of seed inputs.
    \item We design a residual connection mechanism to mitigate the irrelevance as the depth of hop increases during the multi-hop synthesis.
    \item In zero-shot prompting, Mistral-7B fine-tuned with synthetic data from \papertitle~achieves average relative improvements of over $6\%$ and $30\%$ across tasks, compared to Mistral-7B fine-tuned with gold training data and SOTA data synthesis methods. Additionally, \papertitle~enhances the performance of Llama-3.1-8B and Llama-3.2-3B, yielding average relative improvements of approximately $0.7\%$ and $1.5\%$ across tasks, respectively, compared to fine-tuning with gold data.
\end{itemize}

\section{Related Work}
Data synthesis for fine-tuning LLMs targets two primary problems. The first is open-domain generation, which synthesizes data across a wide range of topics and complexity levels. The second is task-specific generation, where synthetic data is tailored to a particular task. One can use the synthetic data in fine-tuning LLMs through techniques, such as instruction tuning, preference tuning, and their variations. This paper focuses on synthesizing training data for instruction tuning to enhance the performance of LLMs for specific tasks. We discuss related methods for data synthesis in both open and task-specific domains in Appendix \ref{related_work}.

Our approach \papertitle~differs from related methods as follows: For each data point, \papertitle~extracts a topic, attributes, and their relationships in the form of knowledge triplets. These triplets then guide the generation of synthetic data relevant to a specific task. \papertitle~also has a residual connection mechanism to maintain the relevance of synthetic data as synthesis depth increases. Additionally, \papertitle~introduces personas to expand attributes, and uses a self-reflection technique to improve diversity and quality of the synthetic task-specific data.

\section{Proposed Method: Attribute-Guided Multi-Hop Data Expansion (AIDE)}
\label{technical_details}
\begin{figure*}[h]
    \centering
     \includegraphics[width=\linewidth]{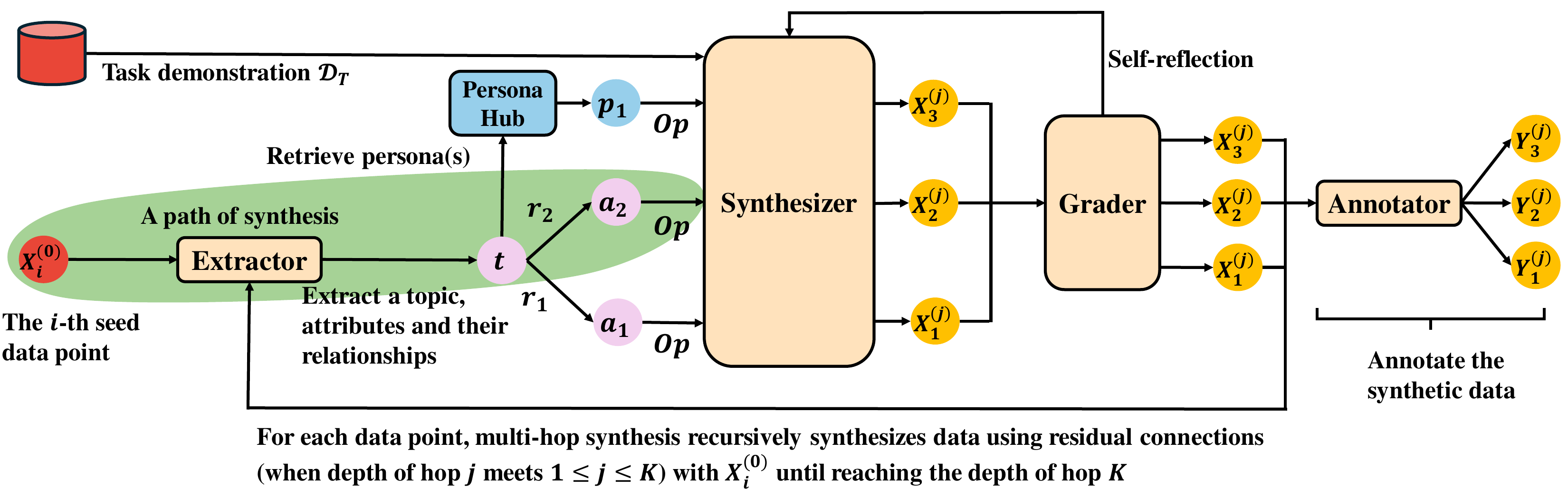}
        \caption{\small Overview of the workflow of \papertitle. $X^{(0)}_i$ denotes the $i$-th task-related seed data point. \papertitle~includes four steps. (1) a LLM extractor extracts a topic $t$, knowledge attributes $\attr_1$ and $\attr_2$ with relationships $r_1$ and $r_2$ of a data point. (2) During the multi-hop synthesis at the depth of hop $j$, a LLM acts as a synthesizer with task demonstrations $\gD_T$ to generate data $X^{(j)}_1$, $X^{(j)}_2$ and $X^{(j)}_3$ along paths of synthesis with a predefined operation $Op$ (i.e., adding constraints). (3) To enhance the diversity of synthesis, we expand attributes by retrieving a persona $p_1$ from a persona hub with $t$. Finally, a LLM as an annotator generates the label of synthetic data. We describe the technical details of \papertitle~in Section \ref{technical_details}.}
\label{overview}
\end{figure*}

\begin{table*}
\small
  \centering
  \setlength{\tabcolsep}{0.7mm}{
  \begin{tabular}{c|c}
  \hline
    Variables & Content\\
    \hline
    \textbf{$X^{(0)}_i$} &  Generate a list of ten items a person might need for a camping trip.\\
    Task demonstration $\gD_T$ & What are the packages pepole needs to prepare for a bike ride through parks or countryside?\\
    \hline
    $<t_1, r_1, \attr_1>$  &  <Outdoor activities, Involves, Camping> \\
    $<t_1, r_2, \attr_2>$  &  <Outdoor activities, Needs, Camping gears> \\
    Persona $p_1$ & An adventurous senior citizen who can recall some related experiences of living in high elevation.\\
    Predefined operation $Op$ & Adding constraint\\
    \hline
    \textbf{$X^{(1)}_1$} & \makecell{What are the top essential items recommended by a survival expert for \\a successful camping trip in harsh weather conditions?}\\
     \hline
    \textbf{$X^{(1)}_2$} & \makecell{Generate a list of ten essential items required for a multi-day camping expedition, \\ensuring that the list includes both shelter and food.} \\
    \hline
     \textbf{$X^{(1)}_3$} &\makecell{Generate a list of ten essential items a person might need for a camping trip, \\ensuring each item is crucial for outdoor activities and aligns with basic camping gear requirements.}\\
     \hline
  \end{tabular}}
  \caption{\small The $1$-hop synthesis in Figure \ref{kg_persona} of Appendix \ref{Unfolded_Multi-Hop} uses an input data point $X^{(0)}_i$ to generate a representation of the data point $\gA^{(0)}_i$ with triplets $<t_1, r_1, \attr_1>$ and $<t_1, r_2, \attr_2>$. We retrieve the persona $P_1$ according to $t_1$. Through the triplets, task demonstrations $\gD_T$, the persona $p_1$ and the predefined operation $Op$, we synthesize $X^{(1)}_1$, $X^{(1)}_2$ and $X^{(1)}_3$ by combining $X^{(0)}_i$ with its corresponding task category and related examples.}
  \label{tab:example_kg}
  \vspace{-4mm}
\end{table*}

In the section, we discuss the details of \papertitle. We define the seed data in a specific task as $D_{\text{seed}}=\{(X_{i}, Y_{i})\}^{n}_{i=1}$ where $n$ is the number of data points in $D_{\text{seed}}$, $X_{i}$ is the $i$-th question and $Y_{i}$ is the corresponding answer to $X_{i}$. We aim to automatically synthesize abundant data within the specific domain by expanding $D_{\text{seed}}$ into $D=\{(X_{i}, Y_{i})\}^{m}_{i=1}$, where $n \ll m$ and $m$ is the size of synthetic dataset. We use the synthetic dataset to fine-tune a model, improving its performance in the specific domain. 

\subsection{Multi-Hop Synthesis}
To synthesize abundant data, we propose a multi-hop synthesis approach, with an example illustrated in Figure \ref{k_synthesis_fig} of Appendix \ref{multi-hop-synthesis}.
\begin{definition}[Multi-hop synthesis]
Given a seed data point $X^{(0)}_i$ where $1\leq i\leq n$, multi-hop synthesis involves recursively generating data from $X^{(0)}_i$ until reaching depth $K$. At depth $K$, $m_K$ denotes the number of $K$-hop neighbors $X^{(K)}$ of $X^{(0)}_i$, where $X^{(K)} = \{X^{(K)}_1, X^{(K)}_2, ..., X^{(K)}_{m_K}\}$. Each $X^{(K)}_i$ for $1\leq i \leq m_K$ is a synthetic data point. The total size of synthetic data after multi-hop synthesis is $m = n(m_1+m_2+...+m_K)$, where $m_1$, $m_2$ and $m_K$ correspond to the number of synthetic data at the depth $1$, $2$, $K$, respectively.
\end{definition}

\subsection{Multi-Hop Synthesis Guided by Attributes and Persona}
During the multi-hop synthesis, we need to ensure the generated data remains relevant to the seed data within the specific task domain. One approach is to use operations as paths in the multi-hop synthesis to create data by rewriting the previous data. However, manually enumerating all possible paths is infeasible, limiting the volume of synthetic data. Furthermore, introducing operations without controlling content along the paths can lead to irrelevant data. To address this, we propose a multi-hop synthesis method guided by attributes and personas, introduced in Sections \ref{syn_kg} and \ref{syn_diversity}, which enhances data diversity while maintaining relevance to the task-related seed data.

\subsubsection{Multi-Hop Synthesis Guided by Attributes for Relevance}
\label{syn_kg}
For a given seed data point, we can extract its main topic, related attributes, and their relationships. Using in-context learning (ICL) \citep{wen2023mindmap, melnyk-etal-2022-knowledge, jin@llmgraph}, a LLM can represent a data point $X^{(K)}_i$ as $\gA^{(K)}_i = \{\langle t, r, \attr \rangle^{(K)}_i|r \in R; t, \attr \in E\}$, where $t$, $r$ and $\attr$ represent the topic, relations and attributes, respectively. $R$ is the set of relations while $E$ contains the topic and attributes. The process of extracting the $\gA^{(K)}_i$ for the $i$-th data $X^{(K)}_i$ is as follows,
\begin{equation}
\label{extract_kg}
    \gA^{(K)}_i = \text{LLM}(X^{(K)}_i).
\end{equation}
We show the prompt of how to extract $ \gA^{(K)}_i$ from $X^{(K)}_i$ in Appendix \ref{app_prompt_kg}. Using $X^{(K-1)}_{i}$ and a triplet $\langle t, r, \attr \rangle^{(K-1)}_i$ from $\gA^{(K-1)}_i$ based on Eq. (\ref{extract_kg}), a LLM synthesizes $X^{(K)}_{i}$ with task demonstrations $\gD_T$. The task demonstrations $\gD_T$ includes task-related examples to guide the process of synthesis. To improve data complexity, we apply operations $Op$ (i.e., adding constraints, reasoning, and concreteness) during synthesis to enhance the quality of synthetic data~\cite{xu2024wizardlm}. This process is summarized as:
\begin{equation}
    X^{(K)}_{i} = \text{LLM}(X^{(K-1)}_{i}, \langle t, r, \attr\rangle^{(K-1)}_i, Op, \gD_T).
\label{syn_eq}
\end{equation}
Prompts for the synthesis process are shown in Appendix \ref{app_prompt_syn}. A multi-hop synthesis example is demonstrated in Figure \ref{kg_persona} in Appendix \ref{Unfolded_Multi-Hop} and Table \ref{tab:example_kg}.

\subsubsection{Multi-Hop Synthesis Guided by Personas for Diversity}
\label{syn_diversity}
\citet{song2024scalingdatadiversityfinetuning} shows that fine-tuning LLMs with diverse data improves performance. However, generating diverse data at scale by LLMs requires varied prompts \citep{chan2024scalingsyntheticdatacreation}. To address this, we leverage Persona Hub \citep{chan2024scalingsyntheticdatacreation} to enhance synthetic data diversity. For each data point, we retrieve the top-$P$ personas by using cosine similarity between its topic embedding and personas embeddings. The retrieved personas $p_i \in P$ guide multi-hop synthesis paths. Given a persona $p_i$, a data point $X^{(K-1)}_i$, task demonstrations $\gD_T$, and a predefined operation $Op$, we synthesize $X^{(K)}_i$ as,
\begin{equation}
    X^{(K)}_i = \text{LLM}(X^{(K-1)}_i, t, p_i, Op, \gD_T).
\end{equation}
Prompts for persona-guided synthesis are shown in Appendix \ref{app_prompt_persona}. Combining multi-hop synthesis with attributes and personas increases the volume of diverse, task-relevant synthetic data.

\subsection{Residual Connection Mechanism for Maintaining Task Relevance} 
\label{residual}
Multi-hop synthesis guided by attributes and personas generates diverse, relevant data, but relevance decreases as hop depth $K$ increases. For instance, synthesizing $10$-hop neighbors introduces unrelated themes (Figure \ref{kg_syn_problem} in Appendix \ref{no_residual_prompt}). To address this drift from the original input at deeper synthesis depths, we introduce residual connections between a seed data point and its neighbors. Specifically, for any depth $d$ where $1 < d \leq K$, we build the connections when $d \leq L$ where $L$ is the depth of residual connection within the range $(1, K]$,
\vspace{-1mm}

{\scriptsize
\begin{flalign*}
X^{(d)}_{i}=&
\begin{cases}
\text{LLM}(X^{(d-1)}_{i}, \langle t, r, \attr \rangle^{(d-1)}_i, Op, \gD_T), & L < d\\
\text{LLM}(X^{(d-1)}_{i}, \langle t, r, \attr \rangle^{(d-1)}_i, Op, \gD_T, X^{(0)}_{i}), & d \leq L.\\
\end{cases}&
\end{flalign*}}We illustrate the detail of residual connection in Appendix \ref{app_residual_connection}. Figure \ref{solve_kg_syn_problem} demonstrates a $10$-hop synthesis using residual connections. Compared to Figure \ref{kg_syn_problem}, the $10$-hop neighbor in Figure \ref{solve_kg_syn_problem} remains focused on the relevant topic.

\section{Experiment}
\begin{table*}[t]
\renewcommand\arraystretch{1.2}
\centering
\resizebox{\linewidth}{!}{
\begin{tabular}{cc|c|ccccc|c|c|c|c|c|c|c}
\hline
\multicolumn{2}{c|}{\multirow{2}{*}{}} &\multirow{2}{*}{\shortstack{Fine-tuning with\\Data Source}} &\multicolumn{5}{c|}{\multirow{1}{*}{MMLU}} & {\multirow{1}{*}{FinBen}} & \multirow{2}{*}{\shortstack{ARC-\\Challenge}} & \multirow{2}{*}{GSM8K} & \multirow{2}{*}{TruthfulQA} & \multirow{2}{*}{MedQA} &\multirow{2}{*}{Avg. ($\uparrow$)}& \multirow{2}{*}{Avg. $\Delta$ ($\uparrow$)}\\
\cmidrule(r){4-9} 

\multicolumn{2}{c|}{}& & Bio. & CS & Phi. & EE & Market. & CFA &&&&&&\\ 
\hline
\multicolumn{2}{c|}{\multirow{1}*{\# Seed Data Points in \textbf{AIDE}}} & & 10 & 10 & 10 & 10 & 10 & 10 & 10 & 10 & 10 & 10 & 10 &\\ 
\hline
\multicolumn{2}{c|}{\multirow{2}*{Pretrained Mistral-7B}}& \textbf{\papertitle~(Ours)} & \textcolor{Orange}{\textbf{75.5\%}}& \textcolor{Orange}{\textbf{57.0\%}} & \textcolor{Orange}{\textbf{72.2\%}} & \textcolor{Orange}{\textbf{60.7\%}} & \textcolor{Orange}{\textbf{89.3\%}}& 41.0\% &  \textcolor{Blue}{\textbf{74.7\%}} & 59.1\% & \textcolor{Orange}{\textbf{69.2\%}} & 44.0\%  & 64.3\% & 7.0\%\\ 
\multicolumn{2}{c|}{\multirow{3}*{}}& Gold training data & 73.2\% & \textcolor{Blue}{\textbf{56.0\%}} & \textcolor{Blue}{\textbf{71.1\%}} & \textcolor{Blue}{\textbf{60.0\%}} & \textcolor{Blue}{\textbf{85.9\%}} & 35.0\% &\textcolor{Orange}{\textbf{79.4\%}} &  53.4\% & 49.9\% & 37.0\% & 60.1\% & NA\\ 
\hline
\multicolumn{2}{c|}{\multirow{2}*{Pretrained Llama-3.1-8B}}& \textbf{\papertitle~(Ours)}& 74.2\% & 47.0\%&	63.0\%&49.7\%&82.1\%& \textcolor{Orange}{\textbf{62.0\%}} & 69.8\% & \textcolor{Blue}{\textbf{65.8\%}} & \textcolor{Orange}{\textbf{69.2\%}} & \textcolor{Orange}{\textbf{56.0\%}} & 63.9\%& 0.7\%\\
\multicolumn{2}{c|}{} & Gold training data & \textcolor{Blue}{\textbf{74.7\%}} & 48.1\% & 60.5\% & 50.1\% & 82.3\% & \textcolor{Blue}{\textbf{61.0\%}} & 69.6\% & \textcolor{Orange}{\textbf{68.2\%}} & 66.1\% & \textcolor{Blue}{\textbf{54.0\%}} & 63.7\%&NA\\
\hline
\multicolumn{2}{c|}{\multirow{2}*{Pretrained Llama-3.2-3B}}& \textbf{\papertitle~(Ours)} & 58.7\% & 43.4\% & 56.6\% & 54.5\% & 71.4\% & 54.0\% & 56.8\% & 45.1\% & \textcolor{Blue}{\textbf{67.6\%}} & 51.0\% & 55.9\% & 1.5\%\\ 
\multicolumn{2}{c|}{} & Gold training data & 60.2\% & 45.0\% & 55.6\% & 48.3\% & 70.7\% & 54.0\% & 56.5\% & 45.5\%& 64.9\% & 50.0\% & 55.1\%&NA\\
\hline
\end{tabular}}
\caption{\small \papertitle-generated data vs. human-curated training data for fine-tuning. We evaluate the performance of various zero-shot learning methods across MMLU, FinBen, ARC-Challenge, GSM8K (8-shot with maj@8), TruthfulQA, and MedQA. We highlight the \textcolor{Orange}{\textbf{best}} and \textcolor{Blue}{\textbf{runner-up}} performances. "Avg." represents the average performance across all benchmarks. For GSM8K, we fine-tune the models using 3.2K gold training data, matching the amount of synthetic data from \papertitle. Results are obtained using the same parameter settings. Avg. $\Delta (\uparrow)$ represents the relative average improvement of models compared to those fine-tuned with gold data. "NA" indicates no difference from models fine-tuned with gold data.}
\label{comparision_gold_training}
\end{table*}

We evaluate \papertitle~to answer the following research questions (RQs): \textbf{(RQ1)} Can \papertitle~enable the fine-tuning of pretrained models that outperform those fine-tuned on human-curated data and data generated by SOTA synthesis methods? \textbf{(RQ2)} How does \papertitle~affect pretrained models' performance under different settings? \textbf{(RQ3)} Does the data from \papertitle~maintain relevance and diversity? 

\subsection{Experiment Setup}
\noindent \textbf{Datasets.} We evaluate all methods across 5 tasks from BIG-Bench, 5 tasks from MMLU, 1 task from FinBen, as well as MedQA, ARC-Challenge, GSM8K, and TruthfulQA. Details of the benchmarks and statistics of the synthetic data from AIDE are provided in Appendix \ref{dataset} and \ref{Stat_syn_data}.

\begin{table}[t]
\Huge
\renewcommand\arraystretch{1.5}
\centering
\resizebox{\linewidth}{!}{
\begin{tabular}{cc|c|ccccc|c}
\hline
\multicolumn{2}{c|}{\multirow{2}{*}{\makecell[c]{Pretrained \\ Model}}} &\multirow{2}{*}{\makecell[c]{Fine-tuning with\\ Data Source}}& \multicolumn{5}{c|}{\multirow{1}{*}{BIG-Bench}}& \multirow{2}{*}{Avg. ($\uparrow$)}\\
\cmidrule(r){4-8}

\multicolumn{2}{c|}{} & & Code & C\&E & Impl. &Math & Time &\\ 
\hline
\multicolumn{2}{c|}{\multirow{5}*{Mistral-7B}}& \textbf{\papertitle~(Ours)} & \textcolor{Orange}{\textbf{91.7\%}}	& \textcolor{Orange}{\textbf{99.2\%}}	& \textcolor{Orange}{\textbf{67.9\%}} & \textcolor{Orange}{\textbf{21.0\%}} & \textcolor{Orange}{\textbf{90.3\%}} & \textcolor{Orange}{\textbf{74.2\%}}\\ 
\multicolumn{2}{c|}{\multirow{4}*{}}& Prompt2Model  &  \textcolor{Blue}{\textbf{84.5\%}}	& 41.2\% 	& 48.0\% & 4.7\% & 2.0\% & 36.1\%\\
\multicolumn{2}{c|}{\multirow{4}*{}}& DataTune  & 73.4\%& 33.8\% & 44.0\% & 8.1\% & 16.9\% & 35.2\%\\
\multicolumn{2}{c|}{\multirow{4}*{}}& Evol-Instruct  & 73.3\%& \textcolor{Blue}{\textbf{73.2\%}} & \textcolor{Blue}{\textbf{65.1\%}} & \textcolor{Blue}{\textbf{14.1\%}} & \textcolor{Blue}{\textbf{45.2\%}} & \textcolor{Blue}{\textbf{54.2\%}}\\ 
\multicolumn{2}{c|}{} & Pretrained Model &46.7\% & 47.7\% & 61.1\% & 11.6\%& 1.4\% & 33.7\%\\
\hline
\end{tabular}}
\caption{\small AIDE vs. SOTA Data Synthesis Methods. We compare the performance of various zero-shot learning approaches in Mistral-7B fine-tuned with AIDE and SOTA synthesis methods across five BIG-Bench tasks. The table follows a setup similar to Table \ref{comparision_gold_training}. Notably, Evol-Instruct fine-tunes Mistral-7B with 250K synthetic data points.}
\label{comparision}
\end{table}

\noindent \textbf{Baselines.} We use fine-tuned Mistral-7B, Llama-3.1-8B, and Llama-3.2-3B with human-generated (gold) data as baselines for comparison with the models fine-tuned using synthetic data from \papertitle. We also compare \papertitle~with SOTA synthesis methods (Evol-Instruct, DataTune, and Prompt2Model) by fine-tuning Mistral-7B. A fine-tuned Mistral-7B using 250K synthetic data from Evol-Instruct \footnote{\url{https:/huggingface.co/dreamgen/WizardLM-2-7B}} is utilized as Mistral-7B with Evol-Instruct. Details about the setups are provided in Appendix \ref{app_experimental_setup}. 

\noindent \textbf{Metrics.} We evaluated all models using zero-shot accuracy as the primary metric on the benchmarks. For GSM8K, we report 8-shot maj@8 performance using prompts from~\citet{wang2023selfconsistency}.

\subsection{Performance and Analysis (RQ1)}
\begin{table}
\tiny
  \centering
\resizebox{1.0\linewidth}{!}{
  \begin{tabular}{ccc|c}
    \hline
    \makecell[c]{Attributes} & Personas & Residual Connections & \makecell[c]{Fine-tuned \\Mistral-7B} \\
    \hline
     \textcolor{Green}{\ding{52}} & \textcolor{red}{\ding{56}} & \textcolor{red}{\ding{56}} & 60.1\% \\
     \textcolor{red}{\ding{56}} & \textcolor{Green}{\ding{52}} & \textcolor{red}{\ding{56}} & 49.3\%\\
      \textcolor{Green}{\ding{52}} & \textcolor{Green}{\ding{52}} & \textcolor{red}{\ding{56}} & 72.2\%\\
    \textcolor{Green}{\ding{52}} & \textcolor{red}{\ding{56}} & \textcolor{Green}{\ding{52}} &75.0\%\\
     \textcolor{Green}{\ding{52}} &\textcolor{Green}{\ding{52}} &\textcolor{Green}{\ding{52}} & \textbf{90.3\%}\\
     \hline
  \end{tabular}}
  \caption{\small Different core components of \papertitle~contribute to the synthetic data, improving the performance of Mistral-7B on the Time task from BIG-Bench. We highlight the \textbf{best} performance and the base performance is in Table \ref{comparision}.}
  \label{tab:core_design}
\end{table}

In Table \ref{comparision_gold_training}, the pretrained models fine-tuned with \papertitle~demonstrate comparable or superior performance to those fine-tuned with gold data. For example, on MMLU tasks, models fine-tuned with \papertitle~data outperform those trained on gold data by an average of $>1.4\%$. In the CFA task, synthetic data from \papertitle~improves Mistral-7B and Llama-3.1-8B by at least $>1.6\%$ compared to gold data. On ARC-Challenge, the Llama series fine-tuned with \papertitle~surpasses their counterparts fine-tuned on gold data. In GSM8K, pretrained models fine-tuned with \papertitle~perform comparably to those fine-tuned with gold data. On TruthfulQA, models fine-tuned with \papertitle~exceed those trained on gold data by an average of $>15.0\%$. Similarly, on MedQA, \papertitle~improves pretrained models by more than $>8.2\%$ on average. In Table \ref{comparision} (BIG-Bench without training sets), Mistral-7B with \papertitle~significantly outperforms itself fine-tuned using Evol-Instruct, Prompt2Model and DataTune by $>20.0\%$, and its pretrained model by $>40.0\%$. This is because Prompt2Model focuses on generating task-specific data with limited diversity, whereas Evol-Instruct, despite its multi-hop synthesis structure, generates data without targeting a specific task.

\begin{figure}[t]
    \centering
    \includegraphics[width=0.65\linewidth]{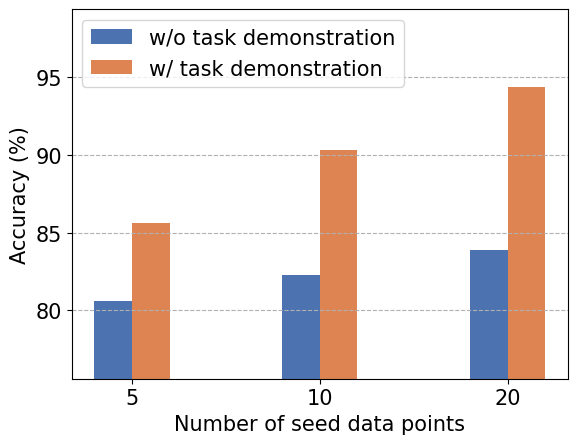}
    \caption{\small The effect of varying the number of seed data w/ and w/o task demonstration on the Time task from BIG-Bench.}
    \label{fig:effect_seed_data}
\end{figure}

\begin{figure}[tbp]
    \begin{subfigure}[t]{0.5\linewidth}
        \centering
        \includegraphics[width=1.\textwidth]{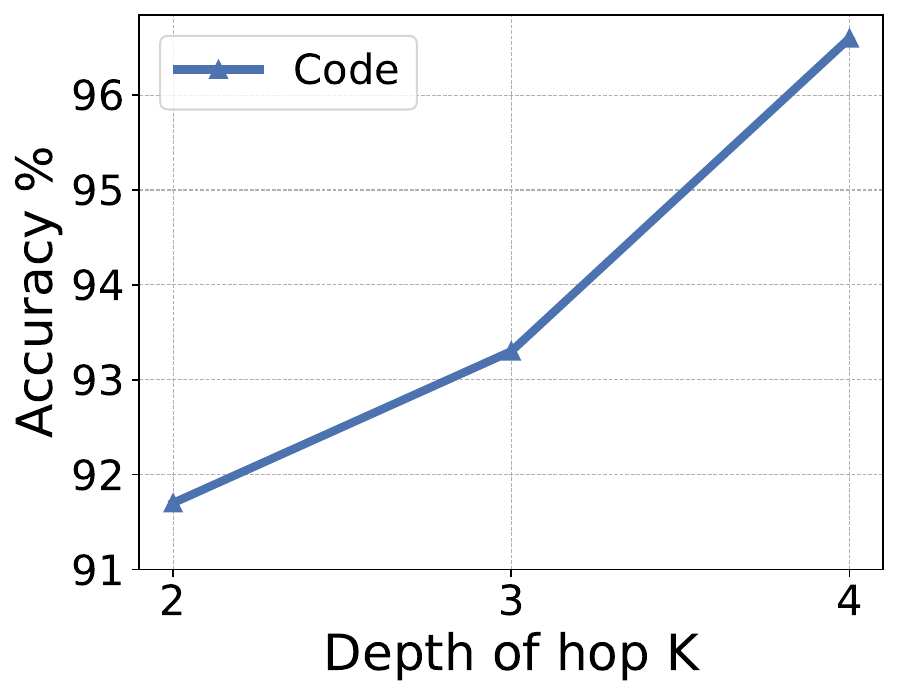}
           \label{K_hop_compare:a}
        \caption{Code}
    \end{subfigure}%
    \begin{subfigure}[t]{0.5\linewidth}
        \centering
        \includegraphics[width=1.\textwidth]{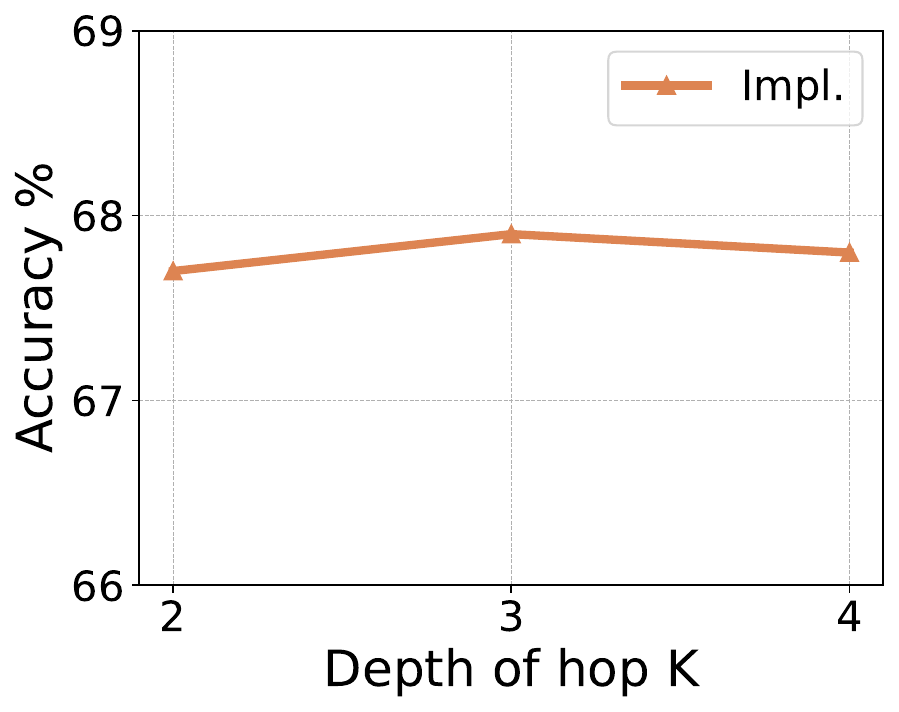}
           \label{K_hop_compare:b}
        \caption{Impl.}
    \end{subfigure}

    \begin{subfigure}[t]{0.5\linewidth}
        \centering
        \includegraphics[width=1.\textwidth]{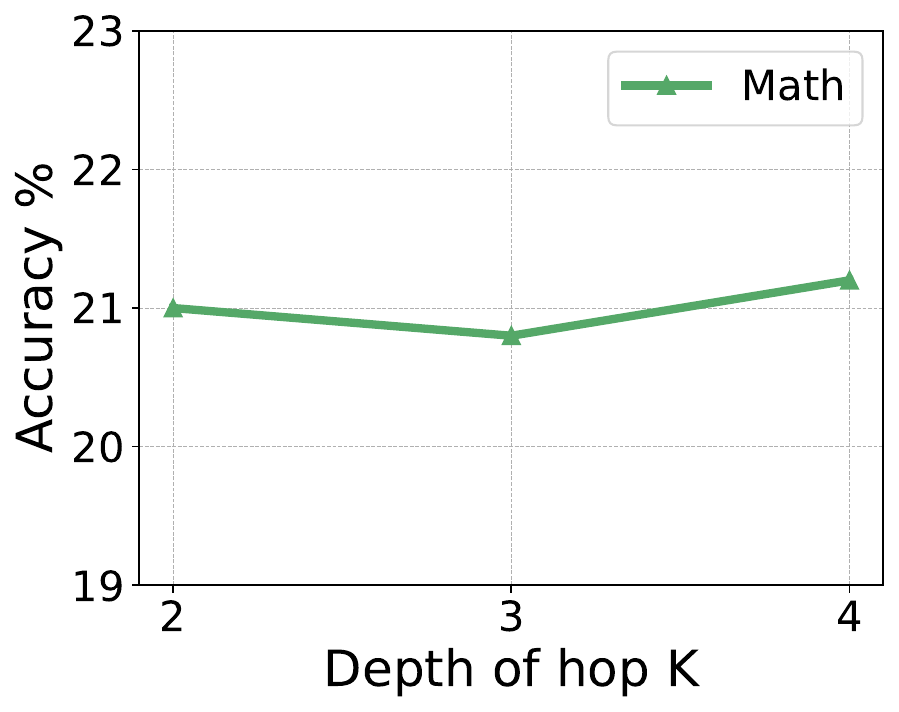}
        \label{K_hop_compare:c}
        \caption{Math}
    \end{subfigure}%
    \begin{subfigure}[t]{0.5\linewidth}
        \centering
        \includegraphics[width=1.\textwidth]{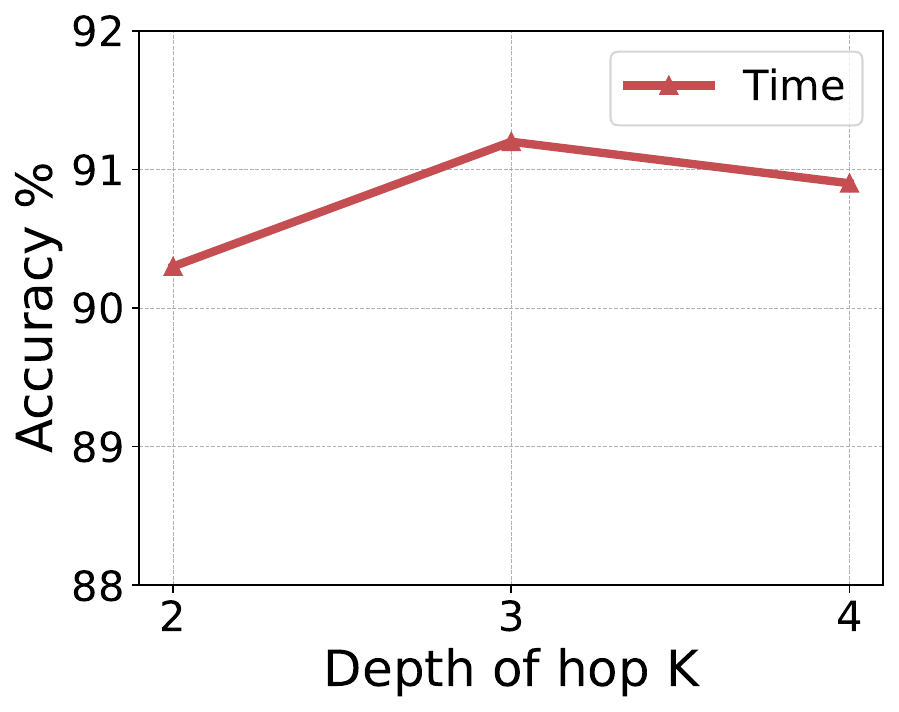}
        \label{K_hop_compare:d}
        \caption{Time}
    \end{subfigure}
\caption{\small Effect of data quantity with different number of $K$ values in multi-hop synthesis based on the BIG-Bench.}
\label{fig:K_quantity}
\end{figure}

\subsection{Ablation and Sensitivity Studies (RQ2)} 
We conduct ablation studies to empirically explore \papertitle~with pretrained models.

\noindent \textbf{Effectiveness of Core Designs.} Table \ref{tab:core_design} (Time task) demonstrates how \papertitle's~core components - attributes, personas, and residual connection - boost Mistral-7B's performance by enhancing the relevance and diversity of synthetic data. To preserve synthesis paths in multi-hop synthesis, we include either attributes or personas. Using only attributes or personas increases Mistral-7B's accuracy from $1.4\%$ to $60.1\%$ and $49.3\%$, respectively. With all three components combined, \papertitle~enables Mistral-7B to achieves $90.3\%$ accuracy, the best performance by preserving synthesis paths and enhancing the relevance of synthetic data.

\noindent \textbf{Effect of Seed Data and Task Demonstration.} The amount of seed data affects initial synthetic data diversity, while task demonstration provide task-related examples to guide synthesis. Therefore, we analyze how the amount of seed data and inclusion of task demonstrations impact \papertitle's~synthetic data quality by fine-tuning Mistral-7B on equal amounts of data. In Figure \ref{fig:effect_seed_data}, we show that increasing seed data in \papertitle~improves Mistral-7B's performance on the Time task through fine-tuning. Furthermore, including task demonstration in \papertitle~boosts Mistral-7B's accuracy by $>10\%$ through fine-tuning, compared to using \papertitle~without task demonstrations.

\noindent \textbf{Scaling with Data Quantity using Different Depth $K$.} The multi-hop depth $K$ determines the amount of \papertitle's~synthetic data, directly influencing fine-tuned model performance. Figure \ref{fig:K_quantity} shows increasing $K$ from 2 to 4  significantly enhances Mistral-7B’s performance on the code task after fine-tuning on \papertitle~data. However, for other tasks, performance gains gradually decrease with higher $K$ values due to the inherent ability gap between the pretrained model and the LLM synthesizer.

\begin{figure}[t]
    \centering
    \includegraphics[width=0.65\linewidth]{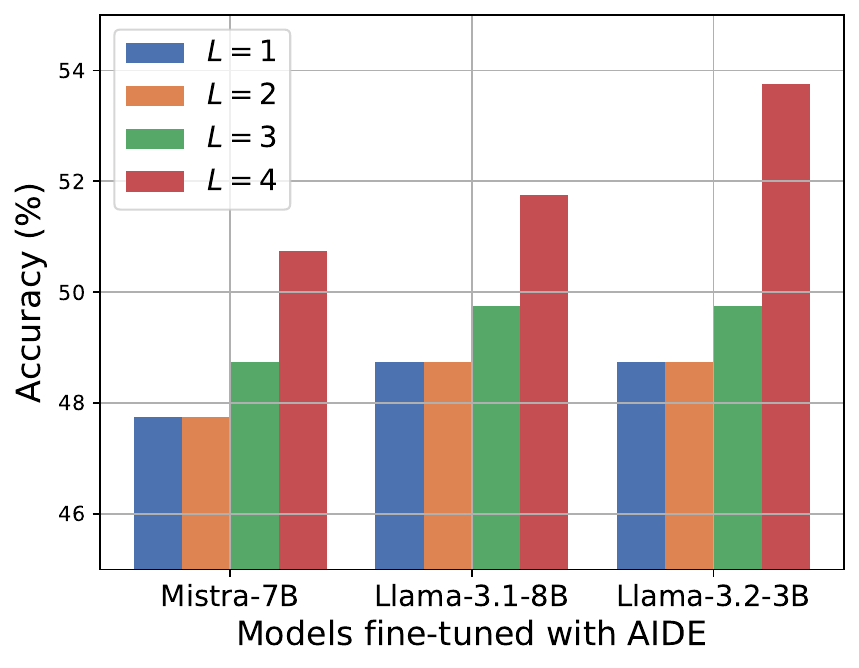}
    \caption{\small The effect of varying the depth of residual connections ($L$) when we fix the hop depth $K$ as 4.}
    \label{fig:effect_residual_connection}
\end{figure}

\noindent \textbf{Effect of Residual Connection.} We use a contract task from LegalBench~\cite{guha2023legalbench}, setting the hop depth $K$ to 4 while varying the depth of residual connections $L$. By synthesizing 5,682 training data points from 6 seeds, we analyze their impact on fine-tuning models. Figure \ref{fig:effect_residual_connection} shows that as the multi-hop synthesis depth increases, a higher residual connection depth $L$ improves the task relevance of the synthetic data, resulting in better model performance during fine-tuning.
\vspace{-1mm}

\noindent \textbf{Effect of Capability of LLMs.} We investigate the impact of using different LLMs as components in \papertitle~by conducting experiments on $5$ BIG-Bench tasks, using Claude Sonnet 3.5 and GPT-3.5-Turbo separately to synthesize data. As shown in Table \ref{llms_sonnet_gpt}, fine-tuning Mistral-7B with \papertitle's~synthetic data, generated with either Claude Sonnet 3.5 or GPT-3.5-Turbo as components, enhances the pretrained model Mistral-7B's performance by $>40.0\%$.
\begin{table}[t]
\Huge
\renewcommand\arraystretch{2}
\centering
\resizebox{\linewidth}{!}{
\begin{tabular}{c|c|c|ccccc|c}
\hline
\multicolumn{2}{c|}{\multirow{2}{*}{Model}} &\multirow{2}{*}{\shortstack{Synthetic \\method}}& \multicolumn{5}{c|}{BIG-Bench Benchmark} & \multirow{2}{*}{Avg.}\\
\cline{4-8}

\multicolumn{2}{c|}{} & & Code & C\&E & Impl. &Math & Time &\\ 
\hline
\multicolumn{2}{c|}{\multirow{3}*{Mistral-7B}}& \makecell[c]{\textbf{\papertitle~(Ours)}\\Claude Sonnet 3.5}  & \textbf{91.7\%}	& \textbf{99.2\%}	& 67.9\% & 21.0\% & \textbf{90.3\%} & 74.0\%\\ 
\cline{3-9}
\multicolumn{2}{c|}{\multirow{3}*{}}& \makecell[c]{\textbf{\papertitle~(Ours)}\\GPT-3.5-Turbo}  & \textbf{91.7\%}& 86.3\%& \textbf{82.5\%} & \textbf{34.6\%} & 85.2\% & \textbf{76.1\%}\\ 
\cline{3-9}
\multicolumn{2}{c|}{\multirow{3}*{}}& -	 &46.7\% & 47.7\% & 61.1\% & 11.6\%& 1.4\% & 33.7\%\\ 
\hline
\end{tabular}}
\caption{\small The performance of Mistral-7B fine-tuned with synthetic data from \papertitle~using different LLMs as synthesizer.}
\label{llms_sonnet_gpt}
      \vspace{-1mm}
\end{table}

\subsection{Relevance and Diversity (RQ3)} 
We empirically investigate the relevance and diversity of synthetic data from \papertitle. Appendix \ref{analysis_visualization} provides details on synthetic data complexity.

\begin{figure}[t]
    \centering
    \includegraphics[width=0.8\linewidth]{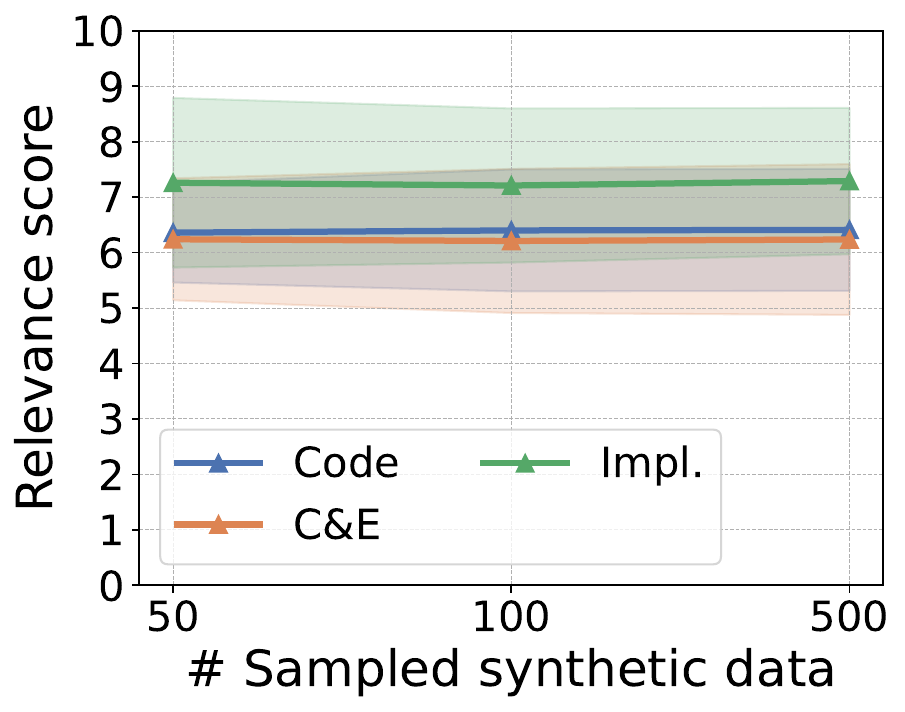}
    \caption{The relevance score related to the sampled synthetic data and task-related seed data from the Code task, the C\&E task and the Impl. task.}
    \label{fig:llm_judge_relevance}
\end{figure}

\begin{figure*}[h]
	\centering
	\begin{subfigure}{0.325\linewidth}
		\centering
		\includegraphics[width=0.9\linewidth]{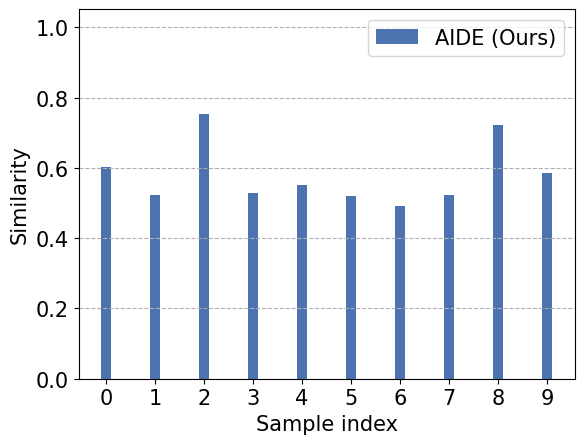}
		\caption{Synthetic data for the Code task}
		\label{relevance_fig:a}
	\end{subfigure}
	\centering
	\begin{subfigure}{0.325\linewidth}
		\centering
		\includegraphics[width=0.9\linewidth]{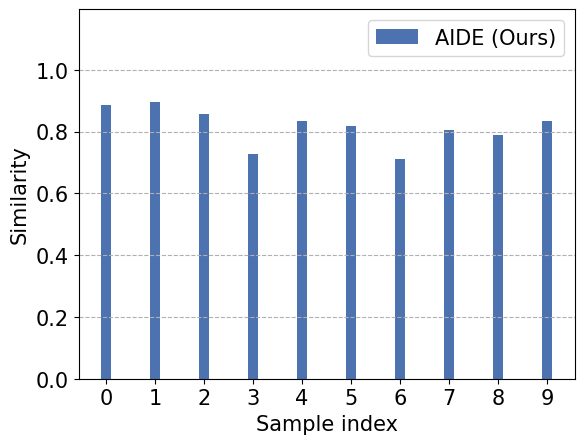}
		\caption{Synthetic data for the C\&E task}
		\label{relevance_fig:b}
	\end{subfigure}
	\centering
	\begin{subfigure}{0.325\linewidth}
		\centering
		\includegraphics[width=0.9\linewidth]{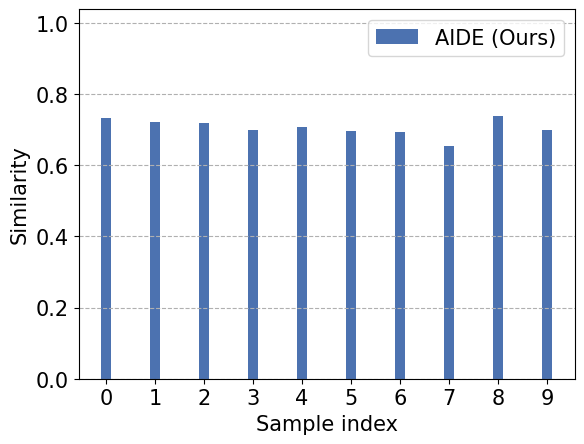}
		\caption{Synthetic data for the Impl. task}
		\label{relevance_fig:c}
	\end{subfigure}
	\caption{For exploring the relevance of synthetic data with the seed data, we compute the similarity between the randomly sampled $10$ synthetic data and the seed data per task. The tasks include Code, Impl. and C\&E.}
	\label{relevance_fig}
\end{figure*}

\begin{table}[t]
\Huge
\renewcommand\arraystretch{1.2}
\centering
\resizebox{\linewidth}{!}{
\begin{tabular}{c|c|c|c}
\hline
Benchmarks & Task Name & \makecell[c]{Diversity of \\ Synthetic Data} (\papertitle)&\makecell[c]{Diversity of \\Gold Data}\\
\hline
\multicolumn{1}{c|}{\multirow{5}*{BIG-Bench}} & Code &  0.59 &\textbf{0.50}  \\ 
\multicolumn{1}{c|}{\multirow{5}*{}}&C\&E &  0.21 &\textbf{0.15}  \\ 
\multicolumn{1}{c|}{\multirow{5}*{}}&Impl.&  0.43  &\textbf{0.40}\\ 
\multicolumn{1}{c|}{\multirow{5}*{}}&Math &\textbf{0.49} &0.50  \\ 
\multicolumn{1}{c|}{\multirow{5}*{}}&Time &  \textbf{0.70}&0.91 \\ 
\hline
\multicolumn{1}{c|}{\multirow{5}*{MMLU}}&Bio. & 0.41& \textbf{0.29} \\
\multicolumn{1}{c|}{\multirow{5}*{}}&CS &  0.66& \textbf{0.24}\\
\multicolumn{1}{c|}{\multirow{5}*{}}&Phi. & 0.49 & \textbf{0.30} \\
\multicolumn{1}{c|}{\multirow{5}*{}}&EE &  0.60&\textbf{0.18}\\
\multicolumn{1}{c|}{\multirow{5}*{}}&Market. & 0.44  &\textbf{0.25} \\
\hline
ARC-Challenge & - &  0.43&\textbf{0.18}\\
\hline
GSM8K & - &  0.43&\textbf{0.21} 	 \\
\hline
TruthfulQA & - & 0.67& \textbf{0.20} \\
\hline
\end{tabular}}
\caption{Quantitative comparison of diversity between synthetic data from \papertitle~for different tasks and gold data from different tasks. We \textbf{highlight lower Self-BLEU scores}, which implies higher diversity.}
\label{comparision2}
\end{table}

\begin{figure*}[htbp]
	\centering
	\begin{subfigure}{0.3\linewidth}
		\centering
		\includegraphics[width=0.9\linewidth]{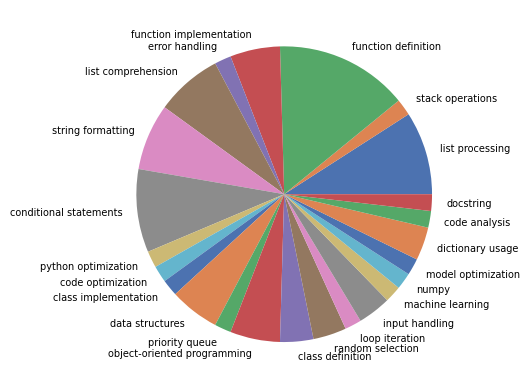}
		\caption{Synthetic data for the Code task}
		\label{diversity_fig:a}
	\end{subfigure}
	\centering
	\begin{subfigure}{0.3\linewidth}
		\centering
		\includegraphics[width=0.9\linewidth]{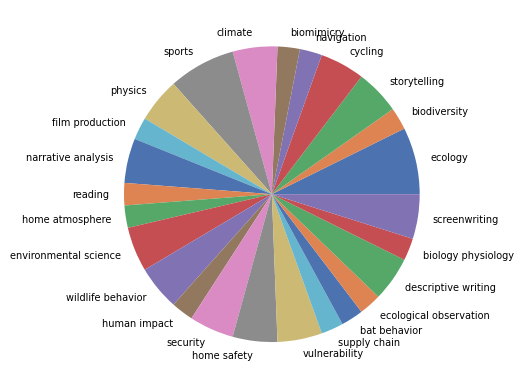}
		\caption{Synthetic data for the C\&E task}
		\label{diversity_fig:b}
	\end{subfigure}
	\centering
	\begin{subfigure}{0.33\linewidth}
		\centering
		\includegraphics[width=0.9\linewidth]{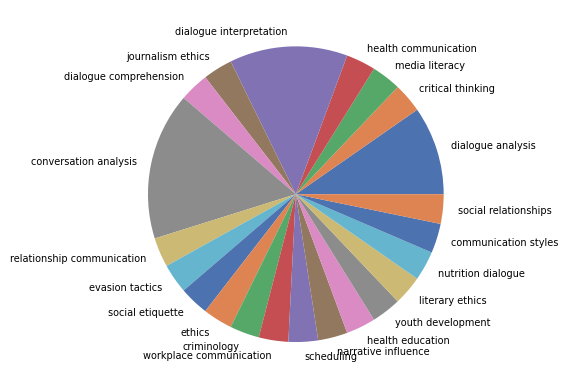}
		\caption{Synthetic data for the Impl. task}
		\label{diversity_fig:c}
	\end{subfigure}
	\caption{We assess the diversity of knowledge by randomly sampling 20 synthetic data points generated by \papertitle~for the Code, C\&E, and Impl. tasks from BIG-Bench.}
	\label{diversity_fig}
\end{figure*}

\noindent \textbf{Analysis of Relevance.} Since the seed data is task-specific, the synthetic data should also be task-relevant if it closely aligns with the seed data. To evaluate this, we randomly sample 10 synthetic data points per task from the Code, C\&E, and Impl. tasks in the BIG-Bench benchmark. We use the Jina embedding model~\cite{günther2023jina} to encode all data points, and compute the similarity between each synthetic data point and its corresponding seed data. As shown in Figure~\ref{relevance_fig}, the synthetic data exhibits strong relevance to the seed data, with an average similarity score above 0.5.

Additionally, we employ Claude Sonnet 3.5 to assess the relevance of synthetic data to the seed data across the three tasks. Claude assigns a relevance score from 0 to 10, with 10 indicating the highest relevance. As shown in Figure~\ref{fig:llm_judge_relevance}, the average scores range from 5 to 9, further confirming the task alignment of the synthetic data. The standard deviation arises because the samples contain data points with significant diversity, yet remain relevant to the corresponding task. 

\noindent \textbf{Analysis of Diversity.} \papertitle~expands attributes through using topics to retrieve personas from Persona Hub, which diversifies the data synthesis. To verify the diversity of synthetic data, we randomly sample $20$ synthetic data per task from the Code, C\&E, and Impl. tasks. Using the prompt shown in Figure\ref{judge_diversity}, we employ Claude Sonnet 3.5 to assess the diversity of the synthetic data based on relevant knowledge. As illustrated in Figure~\ref{diversity_fig:a}, the sampled synthetic data for the Code task covers a variety of programming topics and operations. In the C\&E and Impl. tasks, we observe that the synthetic data spans a wide range of knowledge domains, as shown in Figures~\ref{diversity_fig:b} and~\ref{diversity_fig:c}.

Additionally, following prior work~\cite{ye-etal-2022-zerogen}, we compute Self-BLEU~\cite{10.1145/3209978.3210080} to quantitatively assess the diversity of both synthetic and gold data. The results in Table~\ref{comparision2} show that the synthetic data generated by \papertitle~achieves Self-BLEU scores comparable to those of gold data across most tasks, demonstrating its effectiveness in producing diverse synthetic data.

\section{Conclusion}
Existing data synthesis methods struggle to generate synthetic data that is both task-relevant and diverse for fine-tuning or require large seed datasets. In this paper, we introduce \papertitle, a novel framework that enables task-relevant, diverse, and high-quality data expansion from few seed examples. It features multi-hop synthesis guided by attributes and personas, along with a residual connection to mitigate irrelevance at deeper hops. Our experiments show that fine-tuning Mistral-7B and Llama models with \papertitle~outperforms the models fine-tuned with gold data and SOTA synthesis methods.

\bibliography{custom}

\cleardoublepage
\newpage

\appendix

\section{Detailed Related Work}
\label{related_work}
\noindent \textbf{Data Synthesis for Instruction Tuning in Open Domains.} OpenAI has utilized human annotators to develop diverse instruction-response datasets for training InstructGPT~\citep{NEURIPS2022_b1efde53}. Similarly, Alpaca~\citep{alpaca} and Vicuna~\citep{vicuna2023} explore open-domain instruction tuning using the Llama model. Evol-Instruct~\citep{xu2024wizardlm} offers fine control over instruction complexity, while Tree-Instruct~\cite{zhao-etal-2024-tree} underscores the significance of complexity in LLM alignment. CodecLM~\citep{wang-etal-2024-codeclm} adapts instructions for various tasks. However, these methods lack domain specificity, often introducing irrelevant data. For instance, mixing medical and coding data can negatively impact the fine-tuning process for medical question-answering tasks.

\noindent \textbf{Data Synthesis for Instruction Tuning in Task-specific Domains.} Recent research has focused on generating diverse and relevant datasets through data synthesis. For example, ZeroGen~\cite{ye-etal-2022-zerogen} synthesizes data from task-specific prompts, though challenges arise in domains like multiple-choice, where the label set can be infinite. Methods such as DataTune~\cite{gandhi2024bettersyntheticdataretrieving} and Prompt2Model~\cite{viswanathan-etal-2023-prompt2model} transform existing datasets based on task descriptions, but they rely on large pre-existing collections. Approaches like Self-Guide~\cite{zhao2024selfguide} and ProGen~\cite{ye-etal-2022-progen}, which use limited examples for guiding synthesis, lack sufficient diversity in the generated data.

\section{Multi-Hop Synthesis}
\label{multi-hop-synthesis}
\begin{figure}[hb]
  \includegraphics[width=\linewidth]{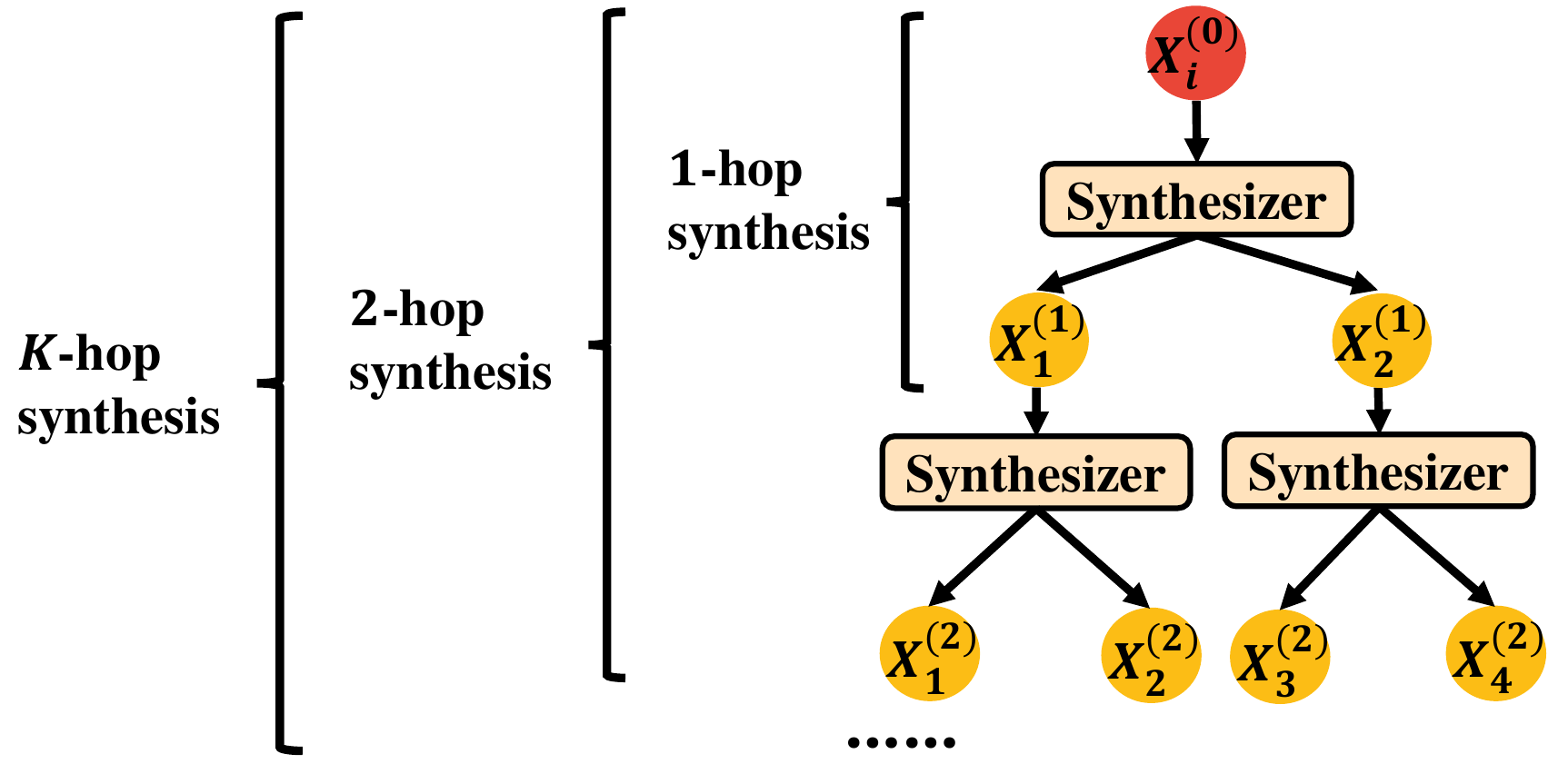}
    \caption{Multi-hop synthesis with the depth of hop $K$ use a seed data point to synthesize new data points. The data points with yellow color represent synthetic data while we use red color to denote a seed data point.}
\label{k_synthesis_fig}
\end{figure}

The Figure \ref{k_synthesis_fig} shows an example of the multi-hop synthesis, which the seed data $X^{(0)}_i$ is used to synthesize its $1$-hop neighbors $X^{(1)}_{1}$ and $X^{(1)}_{2}$ during the $1$-hop synthesis. Similarly, each $1$-hop neighbor can be applied to generate $2$-hop neighbors of $X^{(0)}_i$. For each input data $X^{(0)}_i$ where $1\leq i\leq n$, we recursively synthesis data using the same pattern until reaching the depth of $K$.

\section{An Example of Unfolded Multi-Hop Synthesis }
\label{Unfolded_Multi-Hop}
\begin{figure*}[h]
    \centering
     \includegraphics[width=\linewidth]{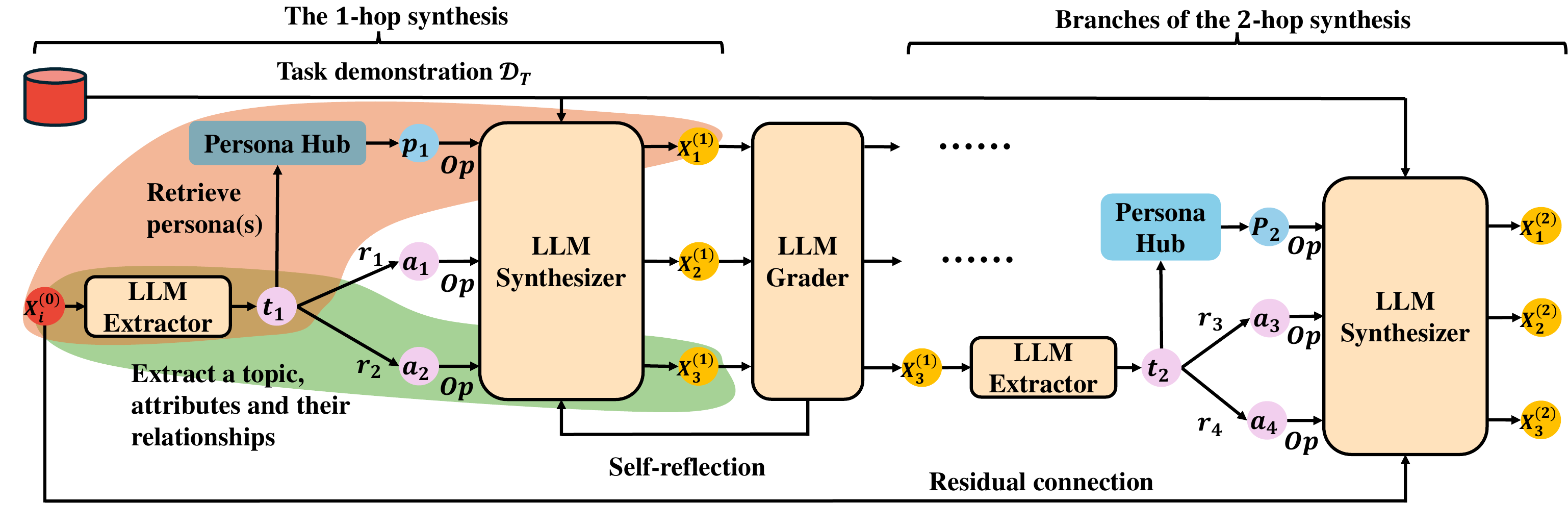}
        \caption{An example of unfolded multi-hop synthesis when $K=2$.}
\label{kg_persona}
\end{figure*}

Figure \ref{kg_persona} illustrates an example of unfolded multi-hop synthesis. In this example, we set $K=2$. $X^{(0)}_i$ is one of the seed data point and $X^{(1)}=\{X^{(1)}_1, X^{(1)}_2, ..., X^{(1)}_{m_1}\}$ represents synthetic data from $1$-hop synthesis while $X^{(2)}=\{X^{(2)}_1, X^{(2)}_2, ..., X^{(2)}_{m_2}\}$ represents synthetic data from $2$-hop synthesis. $r$ is the relation between a topic $t$ and knowledge attribute $\attr$. The predefined operation $Op$ is the abbreviation of operation. Green area includes a path of synthesis showing the relevance between two data points. Orange area shows a path to synthesize data with diversity and relevance. We zoom in one of the branches related to $X^{(1)}_3$ in $2$-hop synthesis. Table \ref{tab:example_kg} demonstrates an example of the synthesis.

\section{Residual Connection}
\label{app_residual_connection} 
We introduce residual connections between a seed data point and its neighbors. Specifically, for any depth $d$ where $1 < d \leq K$, we establish connections when $d \leq L$ where $L$ is the depth of residual connection within the range $(1, K]$. For example, in Figure \ref{kg_persona}, when $K=2$, setting $L=2$ allows connections between the seed data and all neighbors at hop depth 2, ensuring seed information is available for generating the neighbors.

Experiments in Figure \ref{fig:effect_residual_connection} demonstrate that when the hop depth $K$ is large, applying residual connections with a greater depth $L$ enhances the relevance of the synthetic data, leading to improved performance in the fine-tuned model. However, as hop depth $K$ increases, removing low-relevance neighbors instead of using residual connections to retain them can lead to a reduction in the amount of synthetic data.

\section{Detailed Experimental Setup}
\label{app_experimental_setup}
\noindent \textbf{Data Synthesis Setup.} We configure the SOTA data synthesis methods using their default settings. Since BIG-Bench lacks a training set, we sample $10$ task-related seed data points per task from Hugging Face datasets to generate synthetic data. For the remaining benchmarks, we similarly sample $10$ seed data points per task from their respective training sets to produce synthetic data. We set the depth of hop $K=2$ in the multi-hop synthesis. We employ Claude Sonnet 3.5 as the LLM generator, the LLM synthesizer, the LLM grader and the LLM annotator in \papertitle. We require the LLM to generate $\gA^{(K)}$ of a data point $X^{(K)}_i$, which consists of $1$ topic and $3$ most related attributes. Each triplet in $\gA^{(K)}$ followed by $3$ operations: concretizing, adding constraint and adding reasoning. With a topic, we retrieve top-$5$ related personas to diversify attributes.

\noindent \textbf{Fine-tuning Setup.} We applied the LoRA \citep{hu2022lora} to fine-tune Mistral-7B. We randomly split $10\%$ of the synthetic data as validation set while the rest of synthetic data as training set. The process was carried out over $10$ epochs with batch size equal to $10$. We select learning rate $5e-5$ with LoRA's $\alpha$ parameter as $16$ and choose the run with the lowest validation loss at any point. We used the AdamW optimizer \citep{loshchilov2018decoupled} and set LoRA $r = 8$. We conduct our training on a server with $8$ NVIDIA A100 GPUs.

\noindent \textbf{Self-Reflection for Synthetic Data} To ensure the correctness, relevance, and diversity of synthetic data, we apply existing self-reflection techniques \citep{NEURIPS2023_91edff07, pan-etal-2024-automatically} after synthesis (Figure \ref{overview}). A LLM grades synthetic data $X^{(K)}_i$ on these aspects, providing a score (from $1$ to $10$) and feedback. Data exceeding a score threshold (i.e., threshold equal to $5$) is added to the dataset; otherwise, it undergoes limited re-synthesis iterations. A LLM annotator then labels the data, with self-reflection ensuring labeling correctness. Related prompts are shown in Appendix \ref{prompt_self_reflection}.

\section{Statistics of Synthetic Data}
\label{Stat_syn_data}
\begin{table}[t]
\Huge
\renewcommand\arraystretch{1.2}
\centering
\resizebox{\linewidth}{!}{
\begin{tabular}{c|c|c|c|c}
\hline
Benchmarks & Task Name & \makecell[c]{Depth of $K$} &\makecell[c]{Amount of \\ seed Data}& \makecell[c]{Quantity of \\ Synthetic Data}\\
\hline
\multicolumn{1}{c|}{\multirow{5}*{BIG-Bench}} & Code &  2 &10&  3.0K\\ 
\multicolumn{1}{c|}{\multirow{5}*{}}&C\&E &  2 &10&  3.2K\\ 
\multicolumn{1}{c|}{\multirow{5}*{}}&Impl.&  2  &10& 3.1K\\ 
\multicolumn{1}{c|}{\multirow{5}*{}}&Math &2 &10&  3.1K\\ 
\multicolumn{1}{c|}{\multirow{5}*{}}&Time & 2&10& 3.2K\\ 
\hline
\multicolumn{1}{c|}{\multirow{5}*{MMLU}}&Bio. & 2& 10& 3.4K\\
\multicolumn{1}{c|}{\multirow{5}*{}}&CS &  2& 10& 3.2K\\
\multicolumn{1}{c|}{\multirow{5}*{}}&Phi. & 2 & 10& 3.4K\\
\multicolumn{1}{c|}{\multirow{5}*{}}&EE &  2&10& 3.0K\\
\multicolumn{1}{c|}{\multirow{5}*{}}&Market. & 2  &10& 3.3K\\
\hline
ARC-Challenge & - & 2&10& 3.3K\\
\hline
GSM8K & - &  2& 10 & 3.2K\\
\hline
TruthfulQA & - & 2& 10 & 3.1K\\
\hline
\multicolumn{1}{c|}{\multirow{1}*{FinBen}}& CFA & 2  & 10 & 893\\
\hline
MedQA & - & 2& 10 & 2.2K\\
\hline
\end{tabular}}
\caption{Statistics of synthetic data. Note that we adapt the self-reflection mechanism to enhance data quality, which also filters out some synthetic data.}
\label{stat_syn_data_table}
\end{table}

In Table \ref{stat_syn_data_table}, we demonstrate the amount of seed data used and the quantity of data synthesized in \papertitle. Specifically, using $K=2$ and $10$ seed data points for each task, \papertitle~generates approximately $3$K new data points in about $20$ hours when adapting the self-reflection mechanism to improve the quality of new data.

\section{Detailed Analysis of Relevance, Diversity and Complexity (RQ3)}
\label{analysis_visualization}
We conduct experiments to assess whether the synthetic data generated by \papertitle~preserves its complexity.

\subsection{Analysis of Complexity}
\begin{figure*}[htbp]
	\centering
	\begin{subfigure}{0.325\linewidth}
		\centering
		\includegraphics[width=0.9\linewidth]{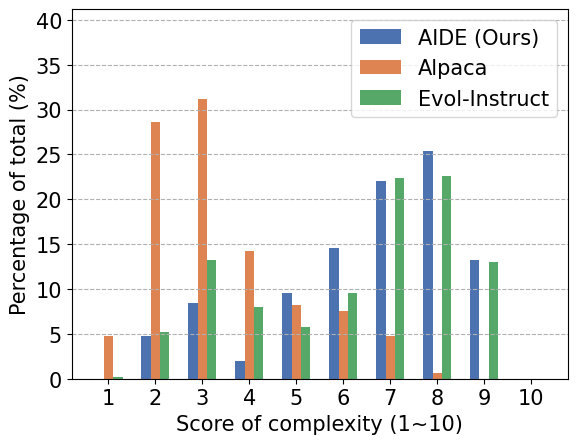}
		\caption{Synthetic data for the Code task}
		\label{complexity_fig:a}
	\end{subfigure}
	\centering
	\begin{subfigure}{0.325\linewidth}
		\centering
		\includegraphics[width=0.9\linewidth]{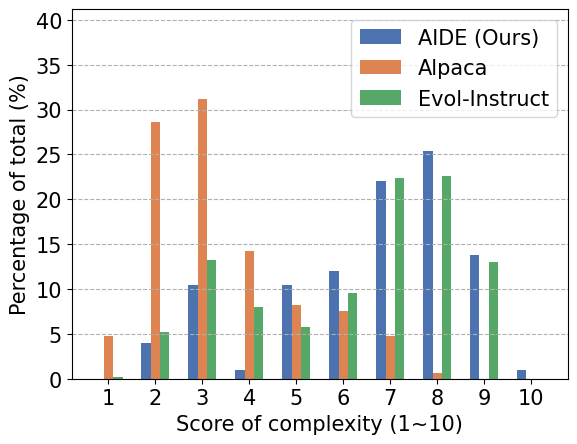}
		\caption{Synthetic data for the C\&E task}
		\label{complexity_fig:b}
	\end{subfigure}
	\centering
	\begin{subfigure}{0.325\linewidth}
		\centering
		\includegraphics[width=0.9\linewidth]{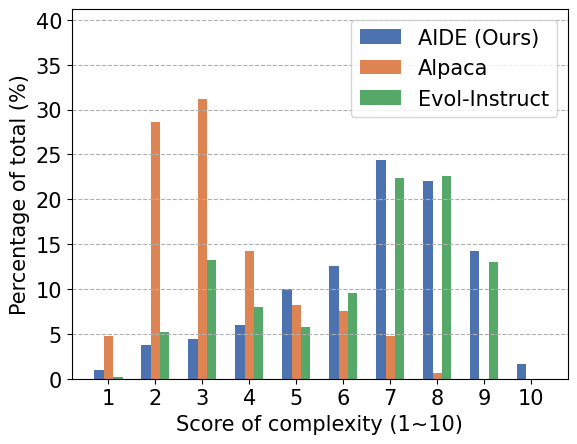}
		\caption{Synthetic data for the Impl. task}
		\label{complexity_fig:c}
	\end{subfigure}
	\caption{The complexity of randomly sampling $500$ synthetic data from \papertitle~based on different domains, including code, cause and effect and implicatures. We also compare the complexity of randomly sampling $500$ synthetic data from the state-of-the-art data synthesis methods including Alpaca and Evol-Instruct.} 
	\label{complexity_fig}
\end{figure*}
 Similar to Evol-Instruct \citep{xu2024wizardlm} using $5$ predefined operations to expand the complexity of synthetic data, \papertitle~utilizes $3$ predefined operations including reasoning, constraint and concrete following triplets from $\gA$ to expand the complexity during data synthesis. For verifying the complexity of synthetic data from \papertitle, we randomly sample $500$ synthetic data from different synthetic methods including Alpaca, Evol-Instructs and our \papertitle. Then we apply Claude Sonnet 3.5 to evaluate the complexity of synthetic data using the same prompt as that from Evol-Instruct. We plot the distribution of score of complexity ranging from $1$ to $10$, shown on Figure \ref{complexity_fig}. We find that most of synthetic data from \papertitle~and Evol-Instruct obtain the score of complexity higher than $5$, when comparing with that from Alpaca. It is worth mentioning that \papertitle~only uses $3$ predefined operations less than the operations applied in Evol-Instruct while having the synthetic data with comparable complexity.
\begin{figure*}[h]
	\centering
	\begin{subfigure}{0.4\linewidth}
		\centering
		\includegraphics[width=1.\linewidth]{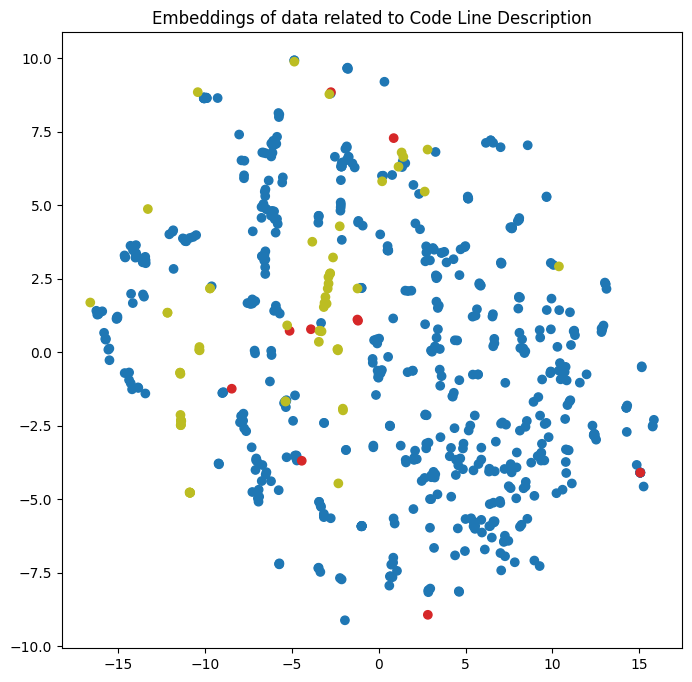}
		\caption{Synthetic data (Code task)}
		\label{tsne_fig:a}
	\end{subfigure}
	\centering
	\begin{subfigure}{0.4\linewidth}
		\centering
		\includegraphics[width=1.\linewidth]{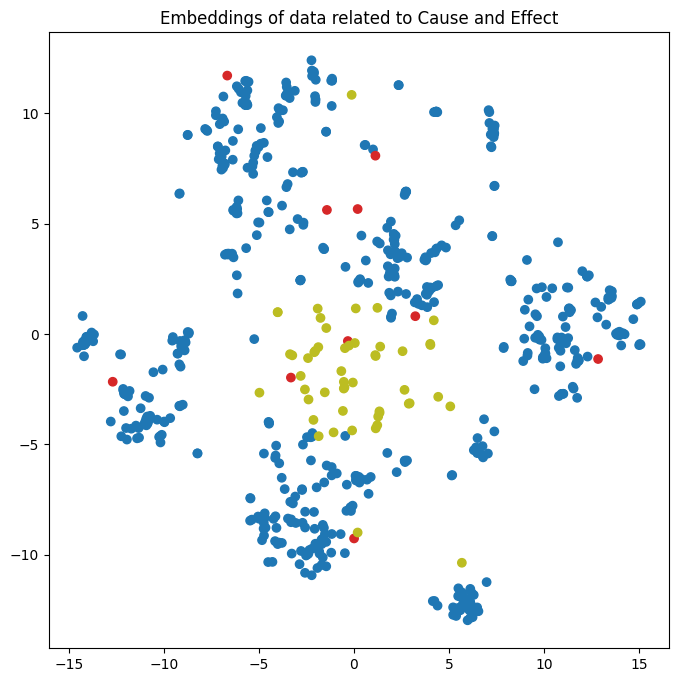}
		\caption{Synthetic data (C\&E task)}
		\label{tsne_fig:b}
	\end{subfigure}
	\caption{We observe that randomly sampling \textcolor{MidnightBlue}{$600$ synthetic data} generated by \papertitle~using the \textcolor{red}{seed data} covers the \textcolor{SpringGreen}{all real test data} from two tasks in the regions of embedding space, after projecting to two dimensions via t-SNE.} 
	\label{tsne_fig}
\end{figure*}

\subsection{Visualization} 
We follow the approach in~\cite{zhao2024wildchat1mchatgptinteraction} and analyze the coverage of synthetic data from \papertitle~in the embedding space. Specifically, we use the jina-embeddings-v2-base-code~\cite{günther2023jina} to embed data points about coding while employ jina-embeddings-v2-base-en to encode other text data. With the embeddings, we utilize t-SNE~\cite{JMLR:v9:vandermaaten08a} to project embeddings into a two-dimensional space. We adopt the real data from the code line description task and the C\&E task as baselines to demonstrate the coverage of synthetic data from \papertitle.

In Figure \ref{tsne_fig:a}, we observe that the embedding clusters of synthetic data via \papertitle~and the embeddings of all real data from the Code task appear to be largely disjoint. Figure \ref{tsne_fig:b} demonstrates that the synthetic data has a larger range which covers all real data from the C\&E task. This supports a conclusion that \papertitle~with few seed data related to specific tasks systematically cover different distributions of the target task space, and therefore fine-tuning Mistral-7B with synthetic data from \papertitle~leads to a positive effect on the improvement of performance of Mistral-7B in specific tasks.

\section{Benchmark Statistics}
\label{dataset}
The details of the benchmarks we employ in the paper are included below:
\begin{itemize}
    \item \textbf{BIG-Bench}~\cite{srivastava2023beyond} includes over $200$ tasks that are currently challenging for language models, encompassing a wide range of categories. We selected the code line description task, cause and effect task, implicatures task, elementary math task and temporal sequence task, totally $5$ tasks, which involve coding understanding, causal reasoning, logical reasoning. The selected tasks without training sets include $60$, $153$, $492$, $7.688$k and $1$k data points in their test sets, respectively.
    \item \textbf{MMLU}~\cite{hendryckstest2021} is designed to evaluate the broad capabilities of language models across $57$ tasks. We select $5$ tasks from the benchmark, including high school biology, college computer science, philosophy, electrical engineering and marketing, which respectively contain $310$, $100$, $311$, $145$ and $234$ data point in the test sets.
    \item \textbf{ARC}~\cite{Clark2018ThinkYH} is a set of grade-school science questions, which are designed to test a model’s ability to perform complex reasoning. We select ARC-Challenge with the more difficult questions that are particularly challenging for AI models because they often require multiple steps of reasoning, inference, and external knowledge beyond the text provided in the question. We apply $1.17$k testing data points in this task to test LLMs.
    \item \textbf{GSM8K}~\cite{cobbe2021gsm8k} is a dataset of $8.5$K high quality linguistically diverse grade school math word problems. The dataset was created to support the task of question answering on basic mathematical problems that require multi-step reasoning. We select the main subset which has $7.47$k training data points and $1.32$k testing data points.
    \item \textbf{TruthfulQA}~\cite{lin-etal-2022-truthfulqa} is a benchmark to measure whether a language model is truthful in generating answers to questions. We select the multiple choice sets which contains $817$ questions for testing. 
    \item \textbf{MedQA}~\cite{jin2020disease} is a comprehensive resource designed to enhance medical question-answering systems. It comprises 10,178 multiple-choice questions sourced from medical exams across the United States, Mainland China, and Taiwan. Each question is accompanied by several answer options, with the correct answer clearly indicated. We select 1,956 data points for the training set and 217 for the validation set. Additionally, we sample 10 seed data points to synthesize 2,173 data points through~\papertitle.
    \item \textbf{FinBen}~\cite{xie2024finbenholisticfinancialbenchmark} is part of the PIXIU project~\cite{xie2023pixiu}, an open-source initiative aimed at developing, fine-tuning, and evaluating large language models (LLMs) in the financial domain. PIXIU encompasses various components including FinBen, a financial language benchmark. The CFA task consists of 1.03k data points, which we divide as follows: 100 data points for the test set, 804 as gold training data, 89 for the validation set, and 10 as seed data points to synthesize 893 additional data points through~\papertitle.
\end{itemize}
\begin{table}[t]
\small
\begin{center}
\begin{tabular}{c|c|c}
\hline
\bf Task Name  & \bf Abbreviation & \bf \# Test data\\
\hline
Code Line Descriptions &   Code     &  60  \\
Cause and Effect & C\&E & 153\\
Implicatures  & Impl.& 492\\
Elementary Math & Math & 7,688\\
Temporal Sequence &  Time & 1,000\\
\hline
High School Biology & Bio. &300\\
College Computer Science & CS &100\\
Philosophy & Phi. &311\\
Electrical Engineering & EE &145\\
Marketing & Market. &234\\
\hline
Flare-cfa & CFA & 100 \\
\hline
ARC-Challenge & - & 1,170\\
\hline
GSM8k &-& 1,320\\
\hline
TruthfulQA &-& 817\\
\hline
MedQA &-& 100 \\
\hline
\end{tabular}
\end{center}
\caption{Data statistic of selected tasks from BIG-Bench, MMLU, ARC-Challenge, GSM8K and Truthful QA.}
\label{sample-table}
\end{table}

\section{Prompt for Extracting a Topic and Knowledge Attributes}
\label{app_prompt_kg}
\begin{figure*}[bp!]
    \centering
     \begin{gblock}{Prompt for extracting a topic and  knowledge attributes of a data point}
I want you to act as an instruction analyzer.\\
\smallskip
Given an instruction, you should recognize its topic and knowledge attributes. You need to list at most 2 knowledge attributes while each knowledge attributes should be transferable and concise with one word or two words. Your should only output the topic within <Topic></Topic> XML tags and output knowledge attributes within <Attributes></Attributes> XML tags. \\
Follow the examples below to analyze <The Given Instruction>\\
\smallskip
<Example>\\
<The Given Instruction> As a sports commentator, describe the winning play in the final seconds of a championship game. </The Given Instruction> \\
<Topic> creative writing </Topic> \\
<Attributes> role-play, sports </Attributes>\\
</Example>\\
\smallskip
\textbf{... Some examples ...}\\
\smallskip
<The Given Instruction> \textcolor{red}{\{Here is instruction.\}} </The Given Instruction>
\end{gblock}
\caption{Prompt for extracting a topic and knowledge attributes.}
\label{extract_topic}
\end{figure*}

We utilize Claude Sonnet 3.5 as the LLM extractor in \papertitle, as shown in Figure \ref{overview}. In Figure \ref{extract_topic}, we demonstrate a prompt used in the LLM extractor to extract a topic and knowledge attributes.

\section{Prompt for Synthesizing Data Points with a Triplet and an Operation}
\label{app_prompt_syn}

We apply Claude Sonnet 3.5 as the LLM synthesizer in \papertitle, as illustrated in Figure \ref{overview}. Figure \ref{syn_triplet} provides an example of a prompt used by the LLM synthesizer to generate a new data point, incorporating a triplet and a constraint operation.

\section{Prompt for Synthesizing Data Points with a Topic and Personas}
\label{app_prompt_persona}
We use Claude Sonnet 3.5 as the LLM synthesizer to generate new data points based on a persona and a constraint operation. Figure \ref{syn_persona} demonstrates a prompt provided to the LLM synthesizer, incorporating both a persona and a constraint operation.

\section{An Example of 10-hop Synthesis without Residual Connection}
\label{no_residual_prompt}
Figure \ref{kg_syn_problem} presents an example of $10$-hop synthesis without applying the residual connection. In multi-hop synthesis, when the hop depth $K$ becomes large (e.g., $K=10$), the synthetic data tends to include more irrelevant information.

\section{An Example of 10-hop Synthesis with Residual Connection}
\label{residual_prompt}
We introduce the residual connection mechanism in \papertitle~, as detailed in Section \ref{residual} and Figure \ref{kg_persona}. Figure \ref{solve_kg_syn_problem} illustrates an example of $10$-hop synthesis incorporating the residual connection.

\section{Prompt for Self-Reflection}
\label{prompt_self_reflection}
During the self-reflection, when multi-hop synthesis synthesizes data through knowledge attributes for maintaining relevance, we apply a LLM as grader to check the relevance of the synthetic data and obtain a relevance score. Similarly, while we generate synthetic data through multi-hop synthesis using persona to expand diversity, a LLM grader checks the diversity of the synthetic data and return a diversity score. We show the prompt about checking relevance and diversity in Figure \ref{check_relevance}. With a self-reflection prompt in Figure \ref{self-reflection}, we collect the score of diversity and relevance as the feedback to process the synthetic data. 

\section{Ethical Considerations}
While \papertitle~is an effective framework for generating diverse, task-relevant data, it's important to consider the ethical implications. With only a few seed data points, \papertitle~leverages LLMs to extract, synthesize, grade, and annotate instruction-response pairs. However, like human annotators, LLMs can occasionally generate unethical, toxic, or misleading content. Although we use self-reflection techniques during synthesis, it’s essential to adopt proven methods for detoxifying and reducing bias in LLM outputs. Stricter inspection and filtering rules should also be applied. Given \papertitle's flexibility, future advances in bias mitigation and fairness can be integrated as additional modules.

\section{Limitations}
We recognize \papertitle~'s limitations in the following two areas, which can serve as inspiration for future research opportunities in the field of data synthesis.

Ethical Consideration. Since our method \papertitle~relies on an LLM to serve as the extractor, synthesizer, grader, and annotator, it may inherit biases and fairness issues from the underlying LLM. However, \papertitle~stands to benefit from improved LLMs that incorporate advanced techniques for reducing bias and enhancing fairness.

Cognitive Process. While \papertitle~helps base models improve their performance in the Math task, the zero-shot performance of the fine-tuned base models remain around 20\%. In the future, a potential future direction is to integrate Chain-of-Thought techniques into \papertitle, such that \papertitle~can provide better synthetic data to enhance reasoning steps of the base models though fine-tuning.
\begin{figure*}[bp!]
    \centering
     \begin{gblock}{Prompt for synthesis with a triplet and a constraint operation}
I want you act as a Prompt Writer. Your objective is to rewrite a given prompt into a more complex instruction to make those famous AI systems (e.g., chatgpt and GPT4) a bit harder to handle. But the rewritten prompt must be reasonable and must be understood and responded by humans. You SHOULD generate the rewritten prompt within <Rewritten Prompt></Rewritten Prompt> XML tags through complicating <The Given Prompt>, such that <Rewritten Prompt> meets the following <EXPECTATIONS>\\
\smallskip
<EXPECTATION 1> The <Rewritten Prompt> SHOULD BE SIMILAR TO \textcolor{red}{\{a seed data point (a residual connection)\}}.\\</EXPECTATION 1>\\
<EXPECTATION 2> The <Rewritten Prompt> can be obtained by adding simple constraints into content in <The Given Prompt>. \\
</EXPECTATION 2>\\
<EXPECTATION 3> The <Rewritten Prompt> is related to \textcolor{red}{\{topic\}} using \textcolor{red}{\{knowledge attribute\}}.\\</EXPECTATION 3> \\
<EXPECTATION 4> Make the <Rewritten Prompt> become as SHORT as possible.\\
</EXPECTATION 4>\\
<EXPECTATION 5> <The Given Prompt>, <Rewritten Prompt>, 'given prompt' and 'rewritten prompt' are not allowed to appear in <Rewritten Prompt>.\\</EXPECTATION 5>\\
\\
Follow the below examples to generate <Rewritten Prompt> by \textcolor{red}{\{adding constraints\}} into <The Given Prompt>.\\
\\
\textbf{... Some examples ...}\\
\\
<The Given Prompt>\textcolor{red}{\{Here is instruction.\}}</The Given Prompt>
\end{gblock}
\caption{Prompt for synthesis with a triplet and an operation}
\label{syn_triplet}
\end{figure*}

\begin{figure*}[bp!]
    \centering
     \begin{gblock}{Prompt for synthesis with a persona and a constraint operation}
     A persona is the aspect of someone's character. You can use the given character to generate a <Created Prompt>. Your goal is to use <The Given Persona> and an operation to create a <Created Prompt> different from <The Given Prompt>.You SHOULD generate the <Created Prompt> through the following actions:\\

<Action 1> <Created Prompt> should related to the topic \textcolor{red}{\{a topic\}}. </Action 1>\\
<Action 2> You need to use <The Given Prompt> and the character from <The Given Persona> to generate <Created Prompt>. </Action 2>\\
<Action 3> <Created Prompt> should be different from <The Given Prompt>. </Action 3>\\
<Action 4> The <Created Prompt> can be obtained by adding simple constraints into the generated content. </Action 4>\\
<Action 5> You should try your best not to make the <Created Prompt> become verbose. </Action 5>\\
Follow the examples below to generate <Created Prompt> and output the <Created Prompt> within <Created Prompt></Created Prompt> XML tags.\\
\\
\textbf{... Some examples ...}\\
\\
<The Given Prompt> \textcolor{red}{\{Here is instruction.\}} </The Given Prompt> \\
<The Given Persona> \textcolor{red}{\{Here is a persona.\}}</The Given Persona>\\
\end{gblock}
\caption{Prompt for synthesis with persona and a constraint operation}
\label{syn_persona}
\end{figure*}

\begin{figure*}
    \centering
     \begin{gblock}{An example of $10$-hop synthesis without the residual connection}
\textbf{Input}: Create an example of \hl{an economic} and \hl{a political issue}.\\
\textbf{$1$-hop neighbor}: Design a scenario that incorporates both \hl{an economic} and \hl{a political quandary}, and introduce a critical factor that significantly influences the outcome.\\
......\\
\textbf{$10$-hop neighbor}: Construct a multifaceted narrative that intricately weaves together various \hl{economic}, \hl{political}, and \hl{sociocultural elements}, entwined with an unexpected turn of events that dramatically shifts the trajectory, demanding a methodical strategy to maneuver through the complex array of repercussions, emphasizing an in-depth exploration of \hl{ethical dilemmas}, alongside an additional dimension probing into the subconscious drivers behind the choices made by each character, all while taking into account the impact of \hl{technological advancements} and how they shape the development of the storyline.
\end{gblock}
\caption{An example of $10$-hop synthesis without the residual connection. When the depth of hop $K$ is large in multi-hop synthesis (i.e., $K=10$), more irrelevant information can be introduced in the synthetic data.}
\label{kg_syn_problem}
\end{figure*}

\begin{figure*}[t]
    \centering
     \begin{gblock}{An example of $10$-hop synthesis with the residual connection}
     \textbf{Input}: Create an example of \hl{an economic} and \hl{a political issue}.\\
     \textbf{$1$-hop neighbor}: Develop a multifaceted scenario encompassing interconnected \hl{economic} and \hl{political challenges}, each influencing the other in a complex and nuanced manner.\\
     ……\\
    \textbf{$10$-hop neighbor}: Craft an engaging narrative interlacing complex \hl{economic} and \hl{political dilemmas}, highlighting their symbiotic nature and profound impact on each other, necessitating a nuanced comprehension of their intricate interdependencies for adept navigation.
     \end{gblock}
        \caption{An example of $10$-hop synthesis with the residual connection shown in Figure \ref{kg_persona}.}
\label{solve_kg_syn_problem}
\end{figure*}

\begin{figure*}[bp!]
    \centering
     \begin{gblock}{Prompt in self-reflection for evaluating the relevance/diversity score of  the synthetic data}
I want you to act as a domain expert to rate the relevance of <The Given Prompt> and <The Original Prompt>.\\
\\
You should give an overall score on a scale of 1 to 10, where a higher score indicates the <The Given Prompt> is more relevant to/different from <The Original Prompt>.\\
You must just give <Score> without any other reasons within the <Score></Score> xml tags.\\
Follow the examples below to analyze and rate relevance of <The Given Instruction> and <The Original Prompt> in <Score>. \\
\\
\textbf{... N Examples ...}\\
\\
Your output should follow the format of examples, which means preserve the same format and show the score within <Score></Score> xml tags.\\
<The Original Prompt> \textcolor{red}{\{Here is the original instruction.\}}  </The Original Prompt>\\
<The Given Prompt> \textcolor{red}{\{Here is the given prompt.\}}  </The Given Prompt>
\end{gblock}
\caption{Prompt in the self-reflection can be used to evaluate the relevance score or diversity score of the synthetic data}
\label{check_relevance}
\end{figure*}

\begin{figure*}[bp!]
    \centering
     \begin{gblock}{Prompt for self-reflection to improve the synthetic data}
I want you to act as a professional data generator.\\
\\
The <Score> from grader shows that the <The Given Prompt> is not  relevant to <Pre-prompt> (or the <The Given Prompt> is highly similar to <Pre-prompt>).\\
You are asked to rewrite <The Given Prompt> as the <Improved Prompt> using the <Pre-prompt>.\\
Generate <Improved Prompt> that improves the <Score> of relevance (or <Score> of diversity) by making <Improved Prompt> relevant to <Pre-prompt> (or by making <Improved Prompt> different from <Pre-prompt>).\\
Must only generate <Improved Prompt> within the <Improved Prompt></Improved Prompt> XML tags. \\
\smallskip
\\
\textbf{... N Examples ...}\\
\\
<Pre-prompt> \textcolor{red}{\{Here is the pre-prompt.\}} </Pre-prompt>\\
<The Given Prompt> \textcolor{red}{\{Here is the given prompt.\}} </The Given Prompt>\\
<Score> \textcolor{red}{\{Here is score.\}} </Score>
\end{gblock}
\caption{Prompt for self-reflection, which can be used to improve the relevance or diversity.}
\label{self-reflection}
\end{figure*}

\begin{figure*}[bp!]
    \centering
     \begin{gblock}{Prompt for a LLM judging the diversity of the synthetic data}
You are a helpful AI assistant for evaluating and rating the difficulty and complexity of the following question.\\

Given an instruction, you should recognize its related knowledge without any explanation.\\
List several most related knowledge, the knowledge should be transferable, so that LLM can leverage them to answer similar questions.\\
Each knowledge should be concise with one word or two words.\\
Follow the examples below to analyze <The Given Instruction>.\\
<Example>\\
<The Given Instruction> As a sports commentator, describe the winning play in the final seconds of a championship game. </The Given Instruction>\\
<Knowledge> sports </Knowledge>\\
</Example>\\
\\
\textbf{... N Examples ...}\\
\\
You must just give the knowledge within the <Knowledge></Knowledge> XML tags without any other reasons.\\
<The Given Instruction> \textcolor{red}{\{Here is the given instruction\}} </The Given Instruction>\
\end{gblock}
\caption{A LLM uses the prompt to judge the diversity of the synthetic data from the perspective of knowledge.}
\label{judge_diversity}
\end{figure*}

\begin{figure*}[bp!]
    \centering
     \begin{gblock}{Prompt for a LLM judging the relevance of the synthetic data}
You are a helpful AI assistant for evaluating and rating the difficulty and complexity of the following question.\\

We would like you to evaluate and rate the relevance of <Instruction1> and <Instruction2> . \\
You should give an overall score on a scale of 1 to 10, where a higher score indicates higher relevance between two instructions. 
You must just give a score without any other reasons.\\
Put the score within the <Score></Score> XML tags.\\
\\
\textbf{... N Examples ...}\\
\\
<Instruction1> \textcolor{red}{\{Here is the Instruction1\}} </Instruction1>\\
<Instruction2> \textcolor{red}{\{Here is the Instruction2\}}  </Instruction2>\\
\end{gblock}
\caption{A LLM uses the prompt to judge the relevance of the synthetic data from the perspective of knowledge.}
\label{judge_relevance}
\end{figure*}

\end{document}